\documentclass{article}

\usepackage{amsmath, amsthm, mathtools, amssymb, color, float, eqnarray, booktabs, makecell, soul, mdframed}
\usepackage[hyphens]{url}
\usepackage[square,numbers]{natbib}
\usepackage[justification=centering]{caption}
\usepackage{todonotes}
\usepackage{comment}
\usepackage{bm}

\pdfoutput=1
\usepackage{subfigure}

\usepackage[most]{tcolorbox}

\newcommand{\B}[1]{\textbf{#1}}
\newcommand{\T}{^{\mbox{\tiny T}}}
\newcommand{\ds}{\displaystyle}
\newcommand{\Part}[2]{\frac{\partial #1}{\partial #2}}
\newcommand{\PartS}[2]{\frac{\partial^2 #1}{\partial #2^2}}

\begin{document}


\title{Extreme Theory of Functional Connections: A Physics-Informed Neural Network Method for Solving Parametric Differential Equations \\
}
\date{}
\author{Enrico Schiassi, Carl Leake, Mario De Florio, Hunter Johnston, Roberto Furfaro, Daniele Mortari}

\maketitle

\begin{abstract}

In this work we present a novel, accurate, and robust physics-informed method for solving problems involving parametric differential equations (DEs) called the \textit{Extreme Theory of Functional Connections} (X-TFC). The proposed method is a synergy of two recently developed frameworks for solving problems involving parametric DEs, 1) the \textit{Theory of Functional Connections} TFC, developed by Mortari et al. \cite{TFC,LDE,NDE}, and 2) the \textit{Physics-Informed Neural Networks} PINN, developed by Raissi et al. \cite{raissi}. Although this paper focuses on the solution of \textit{exact problems} involving parametric DEs (i.e. problems where the modeling error is negligible) with known parameters, X-TFC can also be used for \textit{data-driven solutions} and \textit{data-driven discovery} of parametric DEs. In the proposed method, the latent solution of the parametric DEs is approximated by a TFC constrained expression that uses a Neural Network (NN) as the free-function. This approximate solution form always analytically satisfies the constraints of the DE \cite{TFC}, while maintaining a NN with unconstrained parameters, like the Deep-TFC method by Leake et al. \cite{leake}. X-TFC differs from PINN and Deep-TFC; whereas PINN and Deep-TFC use a deep-NN, X-TFC uses a single-layer NN, or more precisely, an \textit{Extreme Learning Machine} (ELM). This choice is based on the properties of the ELM algorithm developed by Huang et al. \cite{ELM}. In order to numerically validate the method, it was tested over a range of problems including the approximation of solutions to linear and non-linear ordinary DEs (ODEs), systems of ODEs (SODEs), and partial DEs (PDEs). The results show that X-TFC achieves high accuracy with low computational time and thus it is comparable with the other state-of-the-art methods.

\end{abstract}
\section{Introduction}
%
Parametric Differential Equations (DEs) are a powerful tool used for the mathematical modelling of various problems, and are present in fields including, but not limited to, physics, engineering, finance, biology, chemistry, and oceanography. There exist two types of parametric DEs: 1) parametric ordinary DEs (ODEs) which are univariate independent variable equations, and 2) parametric partial DEs (PDEs) which are multivariate independent variable equations. The solution of these equations can be used to simulate, identify, characterize, design, and verify the design of a variety of systems. In many practical problems, it is not trivial to find an analytical solution to these parametric DEs. Thus, for these cases, it is preferred to solve these equations numerically.\\
For the numerical solution of ODEs, a variety of methods exist with the most popular being based on the Runge-Kutta family \cite{RK}. Other methods include finite difference, Chebyshev-Picard iteration \cite{MCPI}, and pseudo-spectral methods \cite{Spectral}. However, a recently developed method called the \textit{Theory of Functional Connections} (TFC), Mortari et al. \cite{TFC,LDE,NDE}, has significantly improved the state-of-the-art for numerically estimating the solutions of parametric ODEs. According to the TFC method, the unknown (or latent \cite{raissi}) solution of the equation is approximated with an expression, called \textit{constrained expression}. The constrained expression is the sum of a function that analytically satisfies the constraints, and a functional containing a freely-chosen function that projects this free-function onto the space of functions that vanish at the constraints. In the classic TFC method, the free-function is chosen to be a linear combination of orthogonal polynomials, such as Legendre or Chebyshev polynomials \cite{LDE,NDE}. While the free-function could easily be defined by an explicit polynomial of a specific degree, orthogonal polynomials are used for their beneficial numerical properties. For example, the properties of the Chebyshev polynomials produce a function that minimizes the maximum error in its application, and therefore, is well suited for approximating other functions \cite{NM_OP,AA}. Studies have shown that the TFC method can be used to numerically estimate parametric linear and non-linear ODEs with machine-level error in milliseconds \cite{LDE, NDE, TFC-Control, FOL}. For this reason, TFC is an appealing choice for many different applications. For example, TFC has already been used to solve ODEs with initial value constraints, boundary value constraints \cite{LDE,NDE}, relative constraints, integral constraints \cite{TFCInt}, and linear combinations of constraints. Additionally, this technique has been utilized to solve optimal control problems including energy optimal \cite{TFC-Control} and fuel optimal landing  trajectories on large planetary bodies \cite{FOL}. Moreover, in De Florio et. al. \cite{rgd}, the TFC was leveraged to solve a class of Rarefied-Gas Dynamics problems, matching the benchmarks published by Barichello et. al \cite{barichello,bari}, and Ganapol \cite{ganapol}.\\
Similarly, Yang et al. in \cite{yang} proposed a method based on Artificial Neural Networks (ANNs), in particular, Legendre Neural Networks (LeNNs). The LeNNs are single layer NNs where the activation functions are Legendre Polynomials and the network is trained via the Extreme Learning Machine (ELM) algorithm, proposed by Huang et al. \cite{ELM}. This algorithm is used for Single-hidden Layer Feed-forward Networks (SLFNs). It randomly selects hidden input weights and biases, and computes the output weights via least-squares \cite{ELM}. Although the results obtained from the LeNN method are fast and accurate, the accuracy is affected (especially in the solution of non-linear ODEs) by the fact that the equation constraints are not analytically satisfied as in the TFC framework. Instead, the method adds the constraints as a penalty to the loss function and are minimized when training the neural network (NN).\\
For PDEs, a multitude of numerical methods exist, the most widely used being the Finite Element Method (FEM) \cite{FEM,FEA1,FEA2,FEA3}. In fact, FEM has been successfully applied to solve PDEs in a variety of fields such as structures, fluids, and acoustics. In this method, the domain is discretized into smaller parts called finite elements and simple approximated equations are used to model these elements. Ultimately, these elements are then assembled into a larger system of equations that model the entire problem. However, the major drawback to FEM is the number of subdivisions needed to capture large variations in the solution. For example, this discretization works well for low-dimensional PDEs, but suffers in higher dimensions; the number of elements grows exponentially with the number of independent variables. Thus, the discretization becomes prohibitive as the number of variables increases. Moreover, in the FEM framework, the PDE is solved at discrete nodes and if the solution is needed at different points, an interpolation scheme is required. As mentioned in Ref. \cite{OrigOdePde}, this affects the accuracy of the solution at the interpolated points. Furthermore, extra numerical techniques are needed to perform further manipulation of the FEM solution such as taking the gradients, or computing the integrals.\\
The solution of PDEs has also become an area of interest in the machine learning community where authors have explored using NNs to solve parametric PDEs. In particular, Raissi et al. \cite{raissi} defined frameworks that use NNs and Deep-NNs (DNNs) to solve these equations as \textit{physics-informed neural networks} (PINNs), wherein the latent solution of a PDE is approximated via a NN or DNN. To solve the equations, the network must be trained to learn the parameters of the approximated latent solution. This would result in a classic regression problem if the training process was not \textit{informed} by the physics governing the equation. To inform the training with the physics governing the problem, the parametric PDE, in its implicit form, is added as a penalty (e.g. a regulator) to the loss function and it is minimized when training the network. Hence, in Raissi et al. \cite{raissi}, the term \textit{physics-informed neural network} was coined.\\ 
The physics-informed methods explored in Refs. \cite{raissi} and \cite{ModernPDE} show that the use of NNs overcomes some of FEM's limitations. One major benefit of physics-informed methods is that the points can be randomly sampled from the domain. Therefore, the discretization of the entire domain into a number of elements that grows exponentially with the scale of the problems is avoided. Moreover, the NN is an analytical approximation of the latent solution. This leads to two major advantages: 1) once the network is trained, no interpolation schemes are needed when estimating the solution at points that did not appear during training, and 2) further solution manipulation, such as computing gradients or integrals, can be done analytically. Although this method was created to numerically estimate the solutions of PDEs, it can also be used to approximate the solutions of ODEs. According to this, both the TFC and the LeNN methods can be seen as physics-informed methods. Demonstrated in Raissi et al.\cite{raissi}, these method can also be used to solve inverse problems for parameter estimation, where the physics phenomena are modeled via parametric DEs. However, what was presented in Refs. \cite{raissi} and \cite{ModernPDE} can still be improved both in terms of solution accuracy and computational speed.\\
The technique mentioned above \cite{raissi,ModernPDE} rely on DNNs to approximate the latent solution. This affects the computational cost as gradient descent based methods are needed to train the networks. This limitation is overcome by expanding the latent solution with a single layer NN trained via the ELM algorithm \cite{pielm} \footnote{In this manuscript single layer NNs trained via the ELM algorithm will be refereed to simply as ELMs}. Another limitation of the DNN approach presented in Refs. \cite{raissi} and \cite{ModernPDE} is that the equation constraints are managed by adding extra terms to the loss function which affects the solution accuracy.\\
One way to overcome this limitation is to approximate the solution in such a way that the constraints must be satisfied, regardless of the values of the training parameters in the network. Lagaris et al. \cite{OrigOdePde} handled constraints in this way when solving both ODEs and PDEs via a physics-informed method that leverages a technique similar to the Coons' patch \cite{CoonsPatch} to satisfy the constraints analytically. As stated in Ref. \cite{leake}, analytical satisfaction of the constraints is of significant interest for a variety of problems. This is particularly true when the confidence in the constraint information is high. Moreover, embedding the constraints in this way allows the NN to sample points only from interior of the domain \cite{leake}. This reduces the number of training points needed, and thereby decreases the computational cost of training. While the method proposed by Lagaris et. al \cite{OrigOdePde} works well for ODEs and low-dimensional PDEs with simple boundary constraints, its application is limited, since it does not provide a generalized framework to derive higher-dimensional or more complex constraints. Fortunately, the TFC framework overcomes this limitation, and thereby extends the work of Lagaris et. al. \cite{OrigOdePde}.\\
In fact, a recent extension of the univariate TFC to $n$-dimensions was formalized in Ref. \cite{M-TFC}.  This extension highlighted a succinct method to derive constrained expressions for value constraints and arbitrary order derivative constraints of $(n-1)$-dimensional manifolds in $n$-dimensional space. This means the TFC framework can generate constrained expressions that satisfy the boundary constraints of multidimensional, large-scale, parametric PDEs \cite{M-TFC-PDE}. In fact, this framework has already been utilized to solve PDEs in combination with machine learning algorithms, such as support vector machines \cite{SVM-TFC} and DNNs \cite{leake}.\\
As previously mentioned, numerical techniques based on the TFC framework to solve ODEs have typically used a linear combination of orthogonal polynomials with unknown coefficients as the free-function. This approach leads to a solution via least-squares for linear problems \cite{LDE}, and via iterative least-squares for nonlinear problems \cite{NDE}. In later studies, this method was extended to the solution of bivariate PDEs \cite{M-TFC-PDE}, where the free function was expressed as the product of two linear combinations of orthogonal polynomials with unknown coefficients. Since the free-function remained linear in these coefficients, a linear least-squares or iterative least-squares method could still be used to estimate the PDEs solution. However, as the dimension of the problem increases or the scale of the problem becomes larger, this process will become computationally prohibitive. As proposed by Leake et al. \cite{leake}, one way to overcome this limitation is to select a NN as the free-function. In this study, the authors used DNNs as the free-function, in a framework called \textit{Deep-TFC}, and the results showed that the method was comparable with the ANN proposed by Lagaris et al. \cite{OrigOdePde} in terms of accuracy and computational cost. While Leake et al. \cite{leake} improved upon what was presented in Lagaris et al. \cite{OrigOdePde} by using the TFC to derive the constrained expressions, both methods rely on gradient based methods to train their networks which are computationally expensive.\\
This article introduces the \textit{Extreme Theory of Functional Connections}, or \textit{Extreme-TFC} (X-TFC), which is a synergy between the Theory of Functional Connections (TFC) and Physics-Informed (PI) methods \cite{raissi}. This is achieved by utilizing the TFC constrained expression and expressing the free-function as an ELM. Therefore, X-TFC is intended to be a novel PI method to solve forward and inverse problems involving parametric DEs. The proposed framework is able to overcome some of the limitations of the TFC approaches and the other state-of-the-art PI methods explained above. Indeed, X-TFC can handle large-scale problems (e.g. PDEs with several independent variables), while preserving high accuracy and low computational time compared to competing methods. Moreover, this method is classified as a PI method for two reasons. First, similar to the Raissi et al. \cite{raissi} PINN method and the PI methods, the X-TFC framework uses a NN to directly solve parametric DEs with known parameters, and for data-driven solutions of parametric DEs with known parameters, where the approximated solution is posed in such a way that the physics of the problems are not violated. Additionally, although in this paper we focus only on the solution of parametric DEs with known parameters (i.e. forward modeling fashion), X-TFC can also be used for data-driven discovery of parametric DEs (i.e. solutions of inverse problems for parameter estimation) both in deterministic and probabilistic (e.g. Bayesian Inversion) fashions.\\
The remainder of this article is organized as follows. First, the theory for the X-TFC framework is explained in detail in Section \ref{sec: method}. Next, the results are presented and discussed in Section \ref{sec: results}. In this section, X-TFC is used to solve a few typical problems of interest in physics and engineering, such as the classic Emden-Fowler equation, Radiative Transfer (RT) equation, and Heat Transfer (HT) equation. Additionally, all results are compared with the other state-of-the-art methods. 

\section{Extreme Theory of Functional Connections}\label{sec: method}

As mentioned previously, the X-TFC framework can be used for solving direct and inverse problems involving parametric DEs with high accuracy and low computational time. When solving direct problems involving parametric DEs, the parameters governing the equations are assumed to be known within a certain accuracy. Therefore, two possible scenarios exist: 1) the parametric DE exactly approximates the physical phenomena that it describes, i.e. there are no modeling errors; hence, these problems are called \textit{exact problems}, and 2) the parametric DE approximates the physical phenomena that it describes with non-negligible modeling errors. In the first scenario no data is needed to solve the equation with high accuracy. Thus, when applying the X-TFC method, the solution of the DE reduces to an unconstrained optimization problem. In the second scenario, data is needed to solve the equation with high accuracy. In this case, the solution of the parametric DE is called a \textit{data-driven solution} \cite{raissi}. In Ref. \cite{raissi}, data-driven solutions of parametric DEs are achieved via training NNs in a supervised fashion that includes the DEs in its implicit form in the cost function as a penalty that guarantees that physics is not violated.

When solving inverse problems involving parametric DEs, the parameters governing the DE are unknown and need to be estimated. These kind of problems are called \textit{data-driven discovery} of parametric DEs \cite{raissi}, as the goal is to discover (i.e. estimate) the parameters that govern the equation by comparing the equation solution with data. For example, a typical field where solving inverse problems is of extreme interest is remote sensing \cite{glam,schiassi,hapke81,hapke96}. For instance, in Ref. \cite{hapke2002}, the authors couple radiative and heat transfer equations to form a set of parametric DEs. The solutions of this set of DEs is compared with real data to discover the thermal inertia and the grain size of planetary regoliths, which are the parameters governing the DEs.

Although this paper focuses on the solution of exact problems, in this section we present the general X-TFC method that can be also used for data-driven solution of parametric DEs, and for data-driven discovery of parametric DEs. We also present in detail how to build the constrained expression \cite{leake}, and give a brief description of the ELM algorithm \cite{ELM}.

\subsection{Method}

As previously stated, parametric DEs are a powerful mathematical tool that can be used to model problems of interest in many different fields such as finance, chemistry, physics, and engineering \cite{leake}. We can express parametric DEs, in their most general implicit form, as,
\begin{equation}\label{eq:parDEgen}
    \gamma f_t + \mathcal{N}\left[f;\bm{\lambda}\right] +\varepsilon -\mathcal{U} = 0
\end{equation}
subject to constraints given by the initial conditions (IC) and boundary conditions (BC). In Equation \eqref{eq:parDEgen}, 
$t\in \mathbb{T} \subseteq \mathbb{R}_{0^+}$, 
$ \bm{x} \in \mathbb{D} \subseteq \mathbb{R}^{n}$, 
$f:=f(t,\bm{x}; \gamma(t,\bm{x}) ,\bm{\lambda}(t,\bm{x}))$ is the unknown (or latent) solution ,
$\gamma:=\gamma(t,\bm{x}) \in \mathbb{G} \subseteq \mathbb{R}$ and $\bm{\lambda}:=\bm{\lambda}(t,\bm{x}) \in \mathbb{L} \subseteq \mathbb{R}^m $ are the parameters governing the parametric DE (which are known when dealing with direct problems, and latent when dealing with inverse problems)
\footnote{
In general, even if it is not explicitly reported in the notation,
$f$ is a function of $t$ and $\bm{x}$, and it is parameterized by $\gamma$ and $\bm{\lambda}$, that in general can be $t$ and $\bm{x}$ dependent as well.
},
$\mathcal{N}\left[ \cdot ; \bm{\lambda} \right]$ is a linear or non-linear operator acting on $f$ and parameterized by $\bm{\lambda}$,
the subscript $t$ refers to the partial derivative of $f$ with respect to $t$, i.e. $f_t = \frac{\partial f}{\partial t}$
\footnote{When dealing with ODEs, the partials derivatives become ordinary derivatives, thus, $f_t =  \frac{\partial f}{\partial t} =\frac{\mathrm{d} f}{\mathrm{d} t}$.},
$\varepsilon$ is the modeling error that is negligible when solving \textit{exact problems},
and $\mathcal{U}$ is a known function that in general can depend on $t$ and $\bm{x}$, and can be parametrized by $\gamma$ and $\bm{\lambda}$.\\
The first step in our general physics-informed framework is to approximate the latent solution $f$ with a constrained expression that analytically satisfies the constraints as follows,
   \begin{equation*}
        f(\B{x}; \bm{ \Theta }) = f_{CE}(\B{x}, g(\B{x}); \bm{\Theta}) = A(\B{x}; \bm{\Theta}) + B(\B{x}, g(\B{x}); \bm{\Theta}),
    \end{equation*}
where $\B{x} = [t,\bm{x}]\T \in \bm{\Omega} \subseteq \mathbb{R}^{n+1}$ with $t \geq 0$, 
$\bm{\Theta} = [\gamma, \bm{\lambda}]\T \in \mathbb{P} \subseteq \mathbb{R}^{m+1}$,
$A (\B{x}; \bm{\Theta})$ analytically satisfies the constraints, and 
 $B(\B{x}, g (\B{x}); \bm{\Theta})$ projects the free-function $g(\B{x})$ onto the space of functions that vanish at the constraints \cite{leake}. 
According to the X-TFC method, the free-function, $g(\B{x})$, is chosen to be a single layer feed forward NN, in particular, an ELM \cite{ELM}. That is,
    \begin{equation*}
    g(\B{x}) = \sum_{j=1}^{L} \beta_j\sigma \left(\B{w}_j\T\B{x} + b_j \right), 
    \end{equation*}
where $L$ is the number of hidden neurons, $\B{w}_j = \left[ w_{j,1},...,w_{j,n+1} \right]\T \in \mathbb{R}^{n+1}$ is the input weights vector connecting the $j^{th}$ hidden neuron and the input nodes, 
$\beta_j \in \mathbb{R}$ with $j=1,...,L$  is the $j^{th}$ output weight connecting the $j^{th}$ hidden neuron and the output node, and $b_j$ is the threshold (aka bias) of the $j^{th}$ hidden neuron,
and $\sigma(\cdot)$ are activation functions.
According to the ELM algorithm \cite{ELM},
input weights and biases are randomly selected and not tuned during the training, thus they are known parameters. The activation functions, $\sigma(\cdot)$, are chosen by the user, so they are also known. Therefore, the only unknowns to compute are the output weights $\boldsymbol{\beta} = \left[ \beta_1,...,\beta_L \right]\T$. Hence we can write,
\begin{equation*}
    f(\B{x}; \bm{\Theta}) = f_{CE}(\B{x}, g(\B{x}); \bm{\Theta}) = f_{CE}(\B{x},\bm{\beta}; \bm{\Theta}).
\end{equation*}
The step-by-step process to derive the constrained expression is provided in Section \ref{sec:CE}. Now that $f$ is approximated with a NN, the second step of the X-TFC physics-informed method is to define a loss function,
\begin{equation*}
    \mathcal{L}(\B{x},\bm{\beta};\bm{\Theta}) = \gamma f_{CE,t} + \mathcal{N}\left[f_{CE};\bm{\lambda}\right] +\varepsilon -\mathcal{U},
\end{equation*}
that according to Raissi et al. \cite{raissi} results in a \textit{physics-informed neural network}. Now, the Mean Square Errors ($MSEs$) for the latent solution, $MSE_f$, and for the \textit{physics-informed neural network}, $MSE_{\mathcal{L}}$, are defined,
\begin{eqnarray*}
    MSE_f &=& \frac{1}{2N}\sum_{p=1}^{N} \left( \widetilde{f}_p - f_{p,CE}(\B{x},\bm{\beta},\bm{\Theta}) \right)^2 \\
    MSE_{\mathcal{L}} &=& \frac{1}{2N}\sum_{p=1}^{N} \left( \mathcal{L}_p(\B{x},\bm{\beta};\bm{\Theta}) \right)^2
\end{eqnarray*}
where $ \left[ \B{x}_p, \widetilde{f}_p \right]_{p=1}^N$ are $N$ distinct sample data points with $\B{x}_p = \left[t_p, x_{p,1},...,x_{p,n}\right]\T \in \mathbb{R}^{n+1}$, and $\widetilde{f}_p \in \mathbb{R}$ for each $p=1,...,N$. The final step is
to compute the latent solution $f$, and in the case of the inverse problem the parameters $\bm{\Theta}$, that minimize the cost function $\mathcal{J}$, 
\begin{equation}\label{eq:cost}
    \underset{\bm{\beta}, \bm{\Theta}}{\min} \; \mathcal{J} = \Gamma_f MSE_f + \Gamma_{\mathcal{L}} MSE_{\mathcal{L}},
\end{equation}
where $\Gamma_f$ and $\Gamma_{\mathcal{L}}$ are positive parameters whose values are chosen based on the relative importance of the MSEs in the computation of the unknowns. In other words, these values are problem-dependent and user specified. Using the X-TFC framework, the problem posed in Eq. \eqref{eq:cost} is an unconstrained regularized regression problem that can be solved via any supervised-learning technique.

As mentioned previously, this article only focuses on the solutions of exact problems; problems in which the parameters $\bm{\Theta}$ are known within a certain accuracy, and the model error, $\varepsilon$, is negligible (i.e. $\varepsilon=0$). Therefore the only quantity to compute is the latent solution, $f$, which is approximated via the constrained expression, $f_{CE}$, as explained above. Thus, the problem reduces to,
\begin{equation}\label{eq:costexact}
    \underset{\bm{\beta}}{\min} \; \mathcal{J} =  \mathcal{L}(\B{x},\bm{\beta};\bm{\Theta}),
\end{equation}
where the loss function, $\mathcal{L}$, of the physics-informed neural network is given by, 
\begin{equation*}
    \mathcal{L}(\B{x},\bm{\beta};\bm{\Theta}) = \gamma f_{CE,t} + \mathcal{N}\left[f_{CE};\bm{\lambda} \right] -\mathcal{U}.
\end{equation*}
The minimization problem given in Eq. (\ref{eq:costexact}) is an unconstrained optimization problem that can be solved via any optimization scheme, such as least-squares (for linear problems \cite{LDE}) or iterative-least-squares (for non-linear problems \cite{NDE}) \cite{leake}.

\subsection{Constrained Expression Derivation}\label{sec:CE}

This section gives a step-by-step derivation of multivariate TFC constrained expressions, which in general are $n+1$ dimensional as they depends on the vector $\B{x}$ as previously defined. As mentioned earlier, the multivariate form of the constrained expression is written as follows,
\begin{align*}
   f_{CE} (\B{x}, g (\B{x})) =& \; \underbrace{\mathcal{M}_{i_1, i_2, \cdots, i_{n+1}}(c(\B{x})) v_{i_1} (t) v_{i_2} (x_1) \cdots v_{i_{n+1}} (x_n)}_{A(\B{x})} + \\ &+ \underbrace{g(\B{x}) - \mathcal{M}_{i_1, i_2, \cdots, i_{n+1}} (g(\B{x})) v_{i_1} (t) v_{i_2} (x_1) \cdots v_{i_{n+1}}(x_n)}_{B(\B{x}, g (\B{x}))}\nonumber
\end{align*}
where $\B{x} = \{t, \; x_1, \; \cdots, \; x_n\}\T$ is a vector of the $n+1$ independent variables, $\mathcal{M}$ is an $(n+1)^{th}$ order tensor containing the boundary conditions $c(\B{x})$, $v_{i_1}, \cdots, v_{i_n+1}$ are vectors whose elements are functions of the independent variables, $g(\B{x})$ is the free-function that can be any function that is defined at the constraints \cite{TFC}. According to Ref. \cite{leake}, $A(\B{x})$ analytically satisfies the constraints and $B(\B{x}, g (\B{x}))$ projects the free-function $g(\B{x})$ onto the space of functions that vanish at the constraints. As already explained, in the X-TFC method, the free-chosen function is an ELM.
A mathematical proof that this form of the constrained expression always analytically satisfies the constraints is given in Ref. \cite{M-TFC}.

Some preliminary mathematical notation is defined here that will assist in the derivation of the constrained expression. For consistency, this article uses the same notation as Ref. \cite{leake}. Let $k\in[1,n+1]$ be the index that refers to the $k^{th}$ independent variable. Let $^kc_q^d := \dfrac{\partial^d c(\B{x})}{\partial x_k^d}\bigg|_{x_k=q}$ be the constraint defined by taking the $d^{th}$ order derivative of the constraint function $c (\B{x})$ and evaluating the result at the $x_k=q$ hyperplane. Moreover, let $^kc_{\B{q}_k}^{\B{d}_k}$ be the vector of $\ell_k$ constraints defined at the $x_k=\B{q}_k$ hyperplanes with derivative orders of $\B{d}_k$, where $\B{q}_k$ and $\B{d}_k \in \mathbb{R}^{\ell_k}$. 
Finally, let $^kb_q^d$ denote the boundary condition operator, where,
\begin{equation*}
    ^kb_q^d[f(\B{x})] = \frac{\partial^d f(\B{x})}{\partial x_k^d}\bigg|_{x_k=q}.
\end{equation*}
This operator takes the $d^{th}$ order derivative with respect to $x_k$ of a function, and then evaluates the result at the $x_k=q$ hyperplane.

Now, the step-by-step process for building the constrained expressions is shown, beginning with the $\mathcal{M}$ tensor. The easiest way to explain this derivation is via an example. As in \cite{leake}, the construction of this $\mathcal{M}$ tensor will be presented via a 3D example with Dirichlet boundary conditions in $x_1$ and initial conditions in $x_2$, and $x_3$ on the domain $x_1,x_2,x_3 \in [0,1]\times[0,1]\times[0,1]$. The $\mathcal{M}$ tensor is constructed in three steps.\\
First, the element $\mathcal{M}_{111} = 0$.
Secondly, the first order sub-tensors of $\mathcal{M}$ specified by keeping one dimension's index free and setting all other dimension's indices to 1 consists of the value $0$ and the boundary conditions for that dimension. That is,
    \begin{equation}\label{eq:M1stOrdersGeneral}
        \mathcal{M}_{1,\dots,1,i_k,1,\dots,1} = \begin{Bmatrix} 0, ^kc_{\B{q}_k}^{\B{d}_k} \end{Bmatrix}.
    \end{equation}
For the example considered here, these first-order sub-tensors are,
    \begin{align*}
        &\mathcal{M}_{i_111} = \big[0,c(0,x_2,x_3),c(1,x_2,x_3)\big]\T \nonumber\\
        &\mathcal{M}_{1i_21} = \big[0,c(x_1,0,x_3),c_{x_2}(x_1,0,x_3)\big]\T\\
        &\mathcal{M}_{11i_3} = \big[0,c(x_1,x_2,0),c_{x_3}(x_1,x_2,0)\big]\T \nonumber.
    \end{align*}
Third, the remaining elements of the $\mathcal{M}$ tensor are those with at least two indices that are not equal to one. These elements are the geometric intersection of the boundary condition elements of the first order tensors given in Eq. (\ref{eq:M1stOrdersGeneral}), plus a sign ($+$ or $-$) that is determined by the number of elements being intersected. In general, this can be formally written as follows,
    \begin{equation*}
            {\cal M}_{i_1 i_2 \dots i_{n+1}} =\, ^1 b^{\B{d}^1_{i_1 - 1}}_{\B{q}^1_{i_1 - 1}} \bigg[\, ^2b^{\B{d}^2_{i_2 - 1}}_{\B{q}^2_{i_2 - 1}} \bigg[ \dots \bigg[\, ^{n+1} b^{\B{d}^{n+1}_{i_n}}_{\B{q}^{n+1}_{i_n}} [c(\B{x})]\bigg] \dots \bigg] \bigg] (-1)^{m+1},
    \end{equation*}
where $m$ is the number of indices for the element that are not equal to one.
Using the example constraints, some of these remaining elements are,
    \begin{align*}
       &M_{133} = -c_{x_2x_3}(x_1,0,0)\\
       &M_{221} = -c(0,0,x_3)\\
       &M_{332} = c_{x_2}(1,0,0).
    \end{align*}
Combining these steps results in the full $\mathcal{M}$ tensor; for the example constraints, the full $\mathcal{M}$ tensor is,
\begin{align*}
    \mathcal{M}_{ij1} &= \begin{bmatrix} 0 & c(0,x_2,x_3) & c(1,x_2,x_3) \\ c(x_1,0,x_3) & -c(0,0,x_3) & -c(1,0,x_3) \\ c_{x_2}(x_1,0,x_3) & -c_{x_2}(0,0,x_3) & -c_{x_2}(1,0,x_3)\end{bmatrix}\\
    \mathcal{M}_{ij2} &= \begin{bmatrix} c(x_1,x_2,0) & -c(0,x_2,0) & -c(1,x_2,0) \\ -c(x_1,0,0) & c(0,0,0) & c(1,0,0) \\ -c_{x_2}(x_1,0,0) & c_{x_2}(0,0,0) & c_{x_2}(1,0,0)\end{bmatrix}\\
    \mathcal{M}_{ij3} &= \begin{bmatrix} c_{x_3}(x_1,x_2,0) & -c_{x_3}(0,x_2,0) & -c_{x_3}(1,x_2,0) \\ -c_{x_3}(x_1,0,0) & c_{x_3}(0,0,0) & c_{x_3}(1,0,0) \\ -c_{x_2x_3}(x_1,0,0) & c_{x_2x_3}(0,0,0) & c_{x_2x_3}(1,0,0)\end{bmatrix}\\
\end{align*}

A standard step-by-step procedure also exists for constructing the $v_{i_k}$ vectors. The general form of $v_{i_k}$ vectors is given by,
\begin{equation*}
    v_{i_k} = \begin{Bmatrix} 1, & \ds\sum_{i=1}^{\ell_k} \alpha_{i1} \, h_i (x_k), & \ds\sum_{i=1}^{\ell_k} \alpha_{i2} \, h_i (x_k), & \dots, & \ds\sum_{i=1}^{\ell_k} \alpha_{i\ell_k} \, h_i (x_k)\end{Bmatrix}\T,
\end{equation*}
where  $h_i (x_k)$ can be any $\ell_k$ linearly independent functions that produce a nonsingular matrix in Eq. \eqref{eq:vVecConst}. 
As suggested in Refs. \cite{TFC,leake,M-TFC},
the simplest set of linearly independent functions are monomials. That is, 
\begin{equation*}
    h_i (x_k) = x_k^{i-1}
\end{equation*}
Finally, the $\ell_k \times \ell_k$ coefficients $\alpha_{ij}$ are simply computed via matrix~inversion,
\begin{equation}\label{eq:vVecConst}
    \begin{bmatrix} ^kb^{d_1}_{q_1}[h_1] & ^kb^{d_1}_{q_1}[h_2] & \dots & ^kb^{d_1}_{q_1}[h_{\ell_k}] \\ ^kb^{d_2}_{q_2}[h_1] & ^kb^{d_2}_{q_2}[h_2] & \dots & ^kb^{d_2}_{q_2}[h_{\ell_k}] \\ \vdots & \vdots & \ddots & \vdots \\ ^kb^{d_{\ell_k}}_{q_{\ell_k}}[h_1] & ^kb^{d_{\ell_k}}_{q_{\ell_k}}[h_2] & \dots & ^kb^{d_{\ell_k}}_{q_{\ell_k}}[h_{\ell_k}] \end{bmatrix} \begin{bmatrix} \alpha_{11} & \alpha_{12} & \dots & \alpha_{1\ell_k} \\ \alpha_{21} & \alpha_{22} & \dots & \alpha_{2\ell_k} \\ \vdots & \vdots & \ddots & \vdots \\ \alpha_{\ell_k 1} & \alpha_{\ell_k 2} & \dots & \alpha_{\ell_k\ell_k} \end{bmatrix} = \begin{bmatrix} 1 & 0 & \dots & 0 \\ 0 & 1 & \dots & 0 \\ \vdots & \vdots & \ddots & \vdots \\ 0 & 0 & \dots & 1\end{bmatrix}.
\end{equation}

\noindent The interested reader can refer to \cite{M-TFC} and \cite{leake} for further details, a mathematical proof that this procedure for generating the $\mathcal{M}$ tensor and the $\B{v}$ vectors produces a valid constrained expression, and for more examples. 

\subsection{Extreme Learning Machine Algorithm}

According to the physics-informed method introduced in this article, the free-chosen function in the TFC constrained expression is chosen to be an ELM. The ELM used in this method is based on the ELM algorithm proposed by Huang et al. \cite{ELM}. ELM is a learning algorithm for Single-hidden Layer Feed-forward Networks (SLFNs) that randomly selects hidden input weights and biases, and computes the output weights via least-squares. That is, input weights and biases are selected randomly and not tuned during the training. Thus, the output weights can be computed by least-squares. Consider $N$ distinct training samples, $ \left[ \B{x}_i, \B{y}_i \right]_{i=1}^N$, where $\B{x}_i = \left[ x_{i1},...,x_{in}\right]\T \in \mathbb{R}^n$ and $\B{y}_i = \left[ y_{i1},...,y_{im}\right]\T \in \mathbb{R}^m$, were a standard SLFN with $L$ hidden neurons and activation function $\sigma(\cdot)$ are used as follows,
\begin{equation*}
    \B{y}_i = \sum_{j=1}^{L} \boldsymbol{\beta}_j\sigma(\B{w}_j\T\B{x}_i + b_j), i=1,...,N
\end{equation*}
where $\B{w}_j = \left[ w_{j1},...,w_{jn} \right]\T \in \mathbb{R}^n$ is the input weight vector connecting the $j^{th}$ hidden neuron and the input nodes, 
$\boldsymbol{\beta}_j = \left[ \beta_{j1},...,\beta_{jm} \right]\T \in \mathbb{R}^m$ is the output weight vector connecting the $j^{th}$ hidden neuron and the output nodes, and $b_j$ is the threshold of the $j^{th}$ hidden neuron. The $N$ equations above can be rewritten in following compact form,
\begin{equation}\label{eq:elmpap}
    \B{H}\B{B}=\B{Y},
\end{equation}
where 
\ $\B{H} \in \mathbb{R}^{N\times L}$ where $H_{ij}= \sigma(\B{w}_j\T\B{x}_i + b_j)$ with $ i=1,...,N$ and $ j=1,...,L$,
$\B{B} \in \mathbb{R}^{L\times m}$ where $\B{B}_i = \boldsymbol{\beta}_i\T$ and $i=1,...,L$,
and $\B{Y} \in \mathbb{R}^{N\times m}$ where $\B{Y}_i = \B{y}_i\T$ and $i=1,...,N$.
As the input weights and biases of the ELM are not tuned, the only unknowns in  Eq. \eqref{eq:elmpap} are $\B{B}$. Thus, \eqref{eq:elmpap} reduces to a least-squares problem. In Ref. \cite{ELM}, $\B{B}$ is computed as follows,
\begin{equation*}
    \B{B}= \B{H}^{\dagger}\B{Y} 
\end{equation*}
where $\B{H}^{\dagger}$ is the Moore-Penrose generalized inverse of the matrix $\B{H}$, which is computed via singular value decomposition (SVD) \cite{ELM}.

The ELM learning algorithm is based on theorem 2.1 and theorem 2.2 of \cite{ELM}. 
These theorems guarantee the existence of the solution of (\ref{eq:elmpap}), for any input weights and bias randomly chosen according to any continuous probability distribution.
In \cite{ELM}, the interested reader can find the formalisation of those theorems and their proofs.

\section{Results} \label{sec: results}

In this section, the X-TFC method is tested on a series of problems of interest in physics and engineering. The problems considered are linear and non-linear ODEs, System of ODEs (SODEs), and PDEs. Moreover, when they are available, the X-TFC method is compared with other state-of-the-art methods and with analytical solutions. The results show that the X-TFC method is as accurate or more accurate than all other methods except the classic TFC; although it should be noted here that in all cases the classic TFC and X-TFC have solutions errors that are on the same order of magnitude. 

\subsection{ODEs}

In this section, X-TFC is applied to linear and non-linear ODEs and SODEs. Each problem was solved in MATLAB on an Intel Core i7 - 9700 CPU PC with 64 GB of RAM.

\begin{figure}[hp]%
\centering
\subfigure[Maximum absolute error as a function of number of points $n$, for fixed number of neurons $L = 100$]{%
\label{fig:ode3anaAccVn}
\includegraphics[width=.75\linewidth]{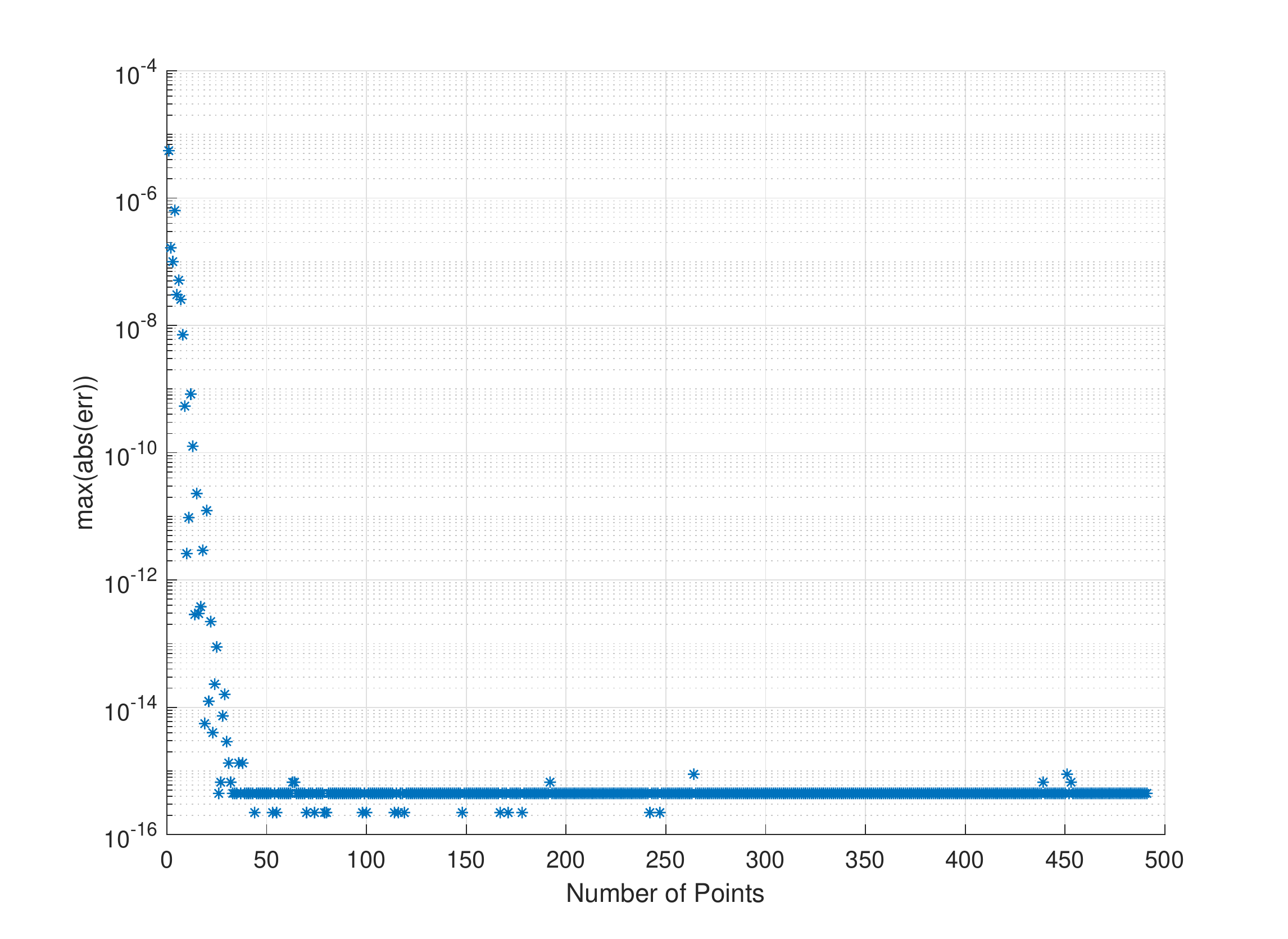}}%
\qquad
\subfigure[Maximum absolute error as a function of number of neurons $L$, for fixed number of points $n = 100$]{%
\label{fig:ode3anaAccVm}
\includegraphics[width=.75\linewidth]{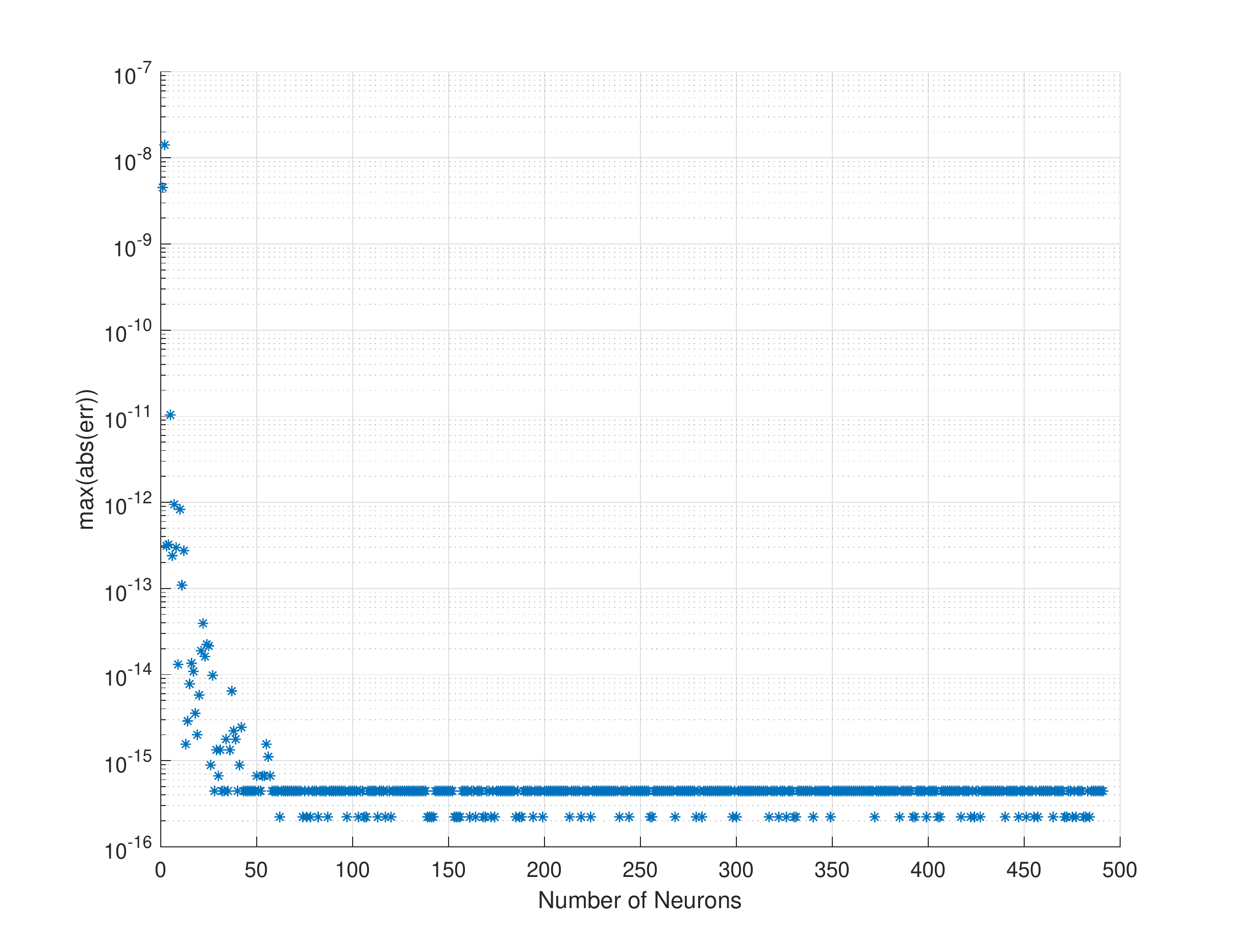}}%
\caption{ Solution accuracy sensitivity analysis for problem \#1 }
\label{fig:ode_analysis}
\end{figure}

Although this manuscript only reports a few examples, the physics-informed X-TFC method was used on several different linear and non-linear ODEs and SODEs, many of which are problems of interest in physics and engineering, such as classic Emden--Folwer equation, advection equation, diffusion equation, advection-diffusion equation, radiative transfer equations, and Bernoulli equations, to name a few. As explained previously, in the X-TFC method there are several hyperparameters that can be modified to obtain accurate solutions. These hyperparameters are the number of training points, $n$, the number of neurons, $L$, the type of activation function, and the probability distribution used to initialize the weights and biases of the ELM. An analysis was performed to study the sensitivity of the X-TFC framework to these hyperparameters. This analysis showed that, for the problems considered, the solution accuracy is not as sensitive to the type of activation function used or to the probability distribution used to initialize the weights and biases as it is to the number of training and the number of neurons. The solution accuracy trends for the number of training points and number of neurons for problem 1 is shown in Figs. \ref{fig:ode3anaAccVn} and \ref{fig:ode3anaAccVm}. Figure \ref{fig:ode3anaAccVn} shows the solution accuracy as a function of the number of training points while holding the number of neurons constant, and Fig.\ref{fig:ode3anaAccVm} shows the solution accuracy as a function of the number neurons for a fixed number of training points. Since the X-TFC methodology uses random numbers to initialize the weights and biases that are untrained, the method is inherently stochastic. Thus, each point in plots of Figs. \ref{fig:ode3anaAccVn} and \ref{fig:ode3anaAccVm} is the maximum absolute error of $10^3$ Monte Carlo simulations. In this article the sensitivity analysis is only shown for problem 1, because the same behaviour was observed for all the problems analyzed.

For all the problems reported in this section, the activation function used in the ELM was a \textit{logistic} activation function, and the weights and biases of the ELM were randomly sampled from a uniform distribution, $(w_j,\,b_j)\,\sim \text{unif}(-10,10)$ where $j=1,...,L$. As previously mentioned, due to the inherently stochastic nature of the X-TFC method, for each problem solved in this section, $10^3$ Monte Carlo simulations were performed to show the variability and test the robustness of the method. 

\subsubsection{Problem 1}
\begin{figure}[hp]
    \centering
    \includegraphics[width=\linewidth]{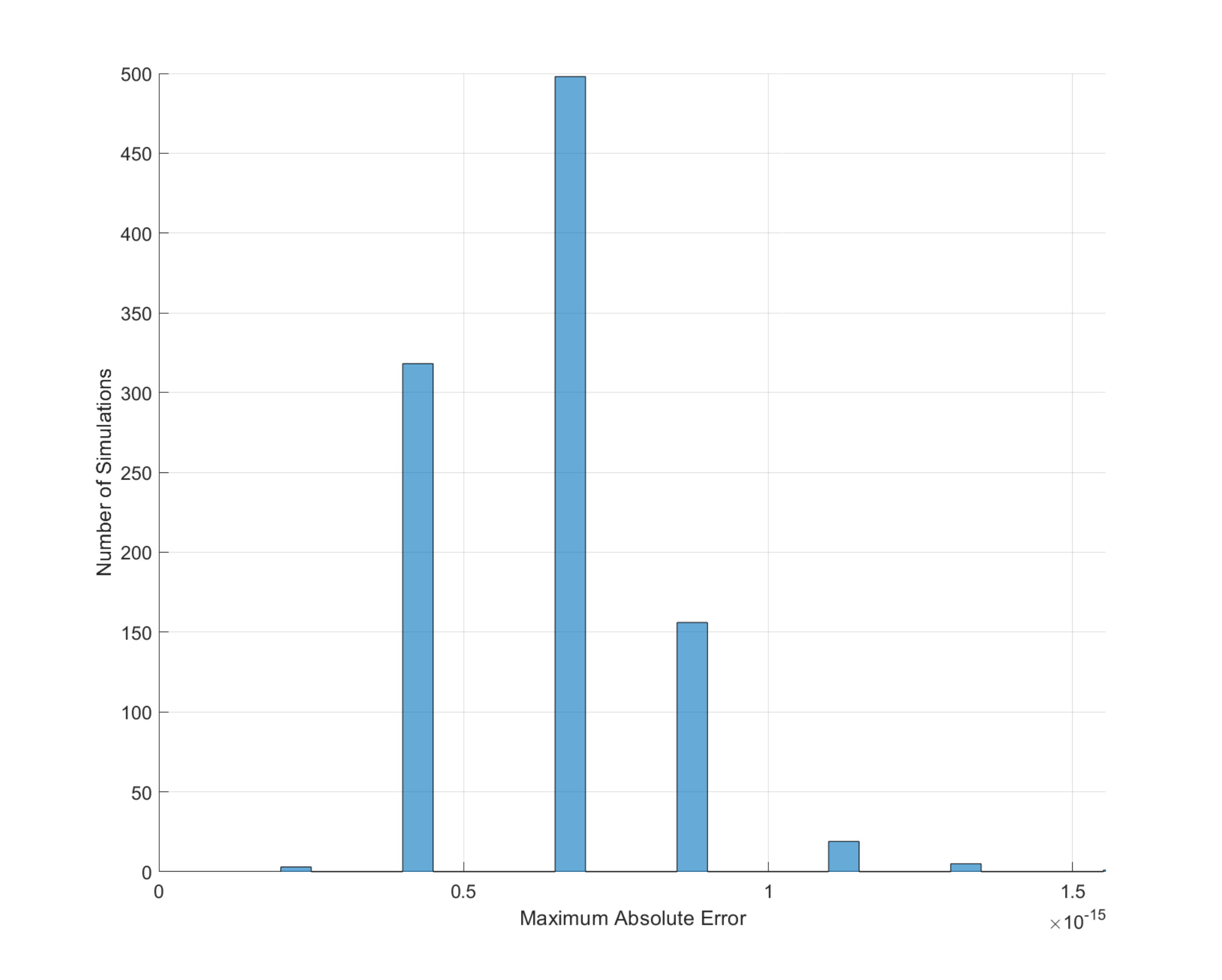}
    \caption{ histogram of $10^3$ Monte Carlo simulations for problem 1 with a \textit{logistic} activation function, $n=50$, and $L=50$.}
    \label{fig:ode3mc}
\end{figure}

\begin{figure}[hp]
    \centering
    \includegraphics[width=\linewidth]{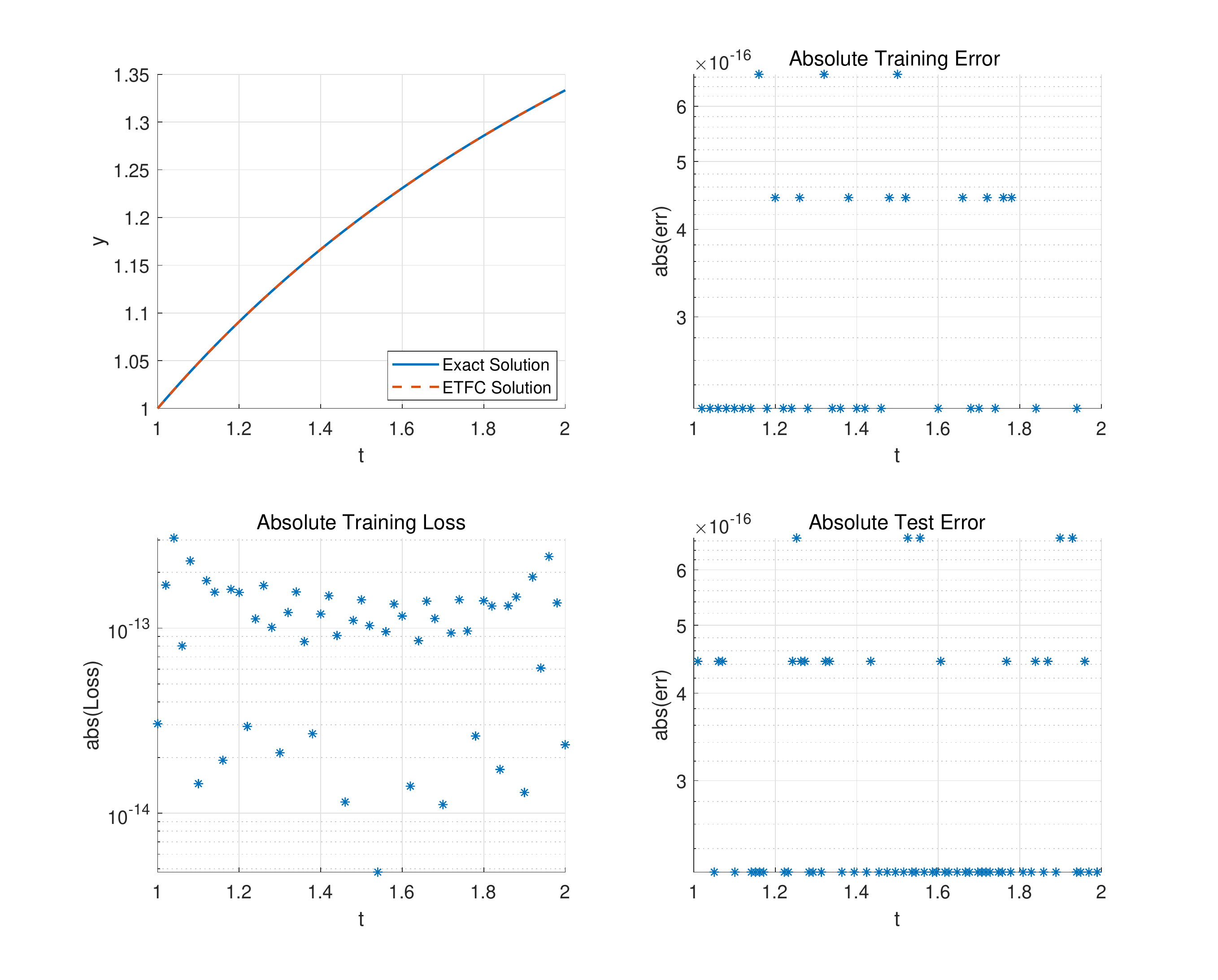}
    \caption{Solution and performances for problem 1: exact solution versus X-TFC solution, absolute training error, absolute loss, and absolute test error, with a \textit{logistic} activation function, $n=51$, and $L=51$. The convergence is achieved in $5$ iterations with a tolerance set to $ 4.440892098500626 \times10^{-16} $. The computational time is $ 1.0993 $ milliseconds, the maximum absolute training error is $ 6.66134 \times10^{-16} $, the maximum absolute training loss is $ 3.0681\times10^{-13} $, and the maximum absolute test error is $ 6.66134\times10^{-16} $}
    \label{fig:ode3}
\end{figure}

The following equation is a non-linear ODE taken from Yang et al. \cite{yang},
\begin{equation*}
    y_{tt} = \frac{1}{2 \, x^2} (y^3 - 2 y^2)
\end{equation*}
subject to $y(1)=1$ and $y(2) = 4/3$ for $t\in[1,2]$. The exact solution is $y (t) = \dfrac{2 t}{t + 1}$. The constrained expression for this problem is,
\begin{equation*}
    y(t) = \left[  \bm{\sigma} + (t-2) \bm{\sigma}_0 + (1-t) \bm{\sigma}_f   \right]\T \bm{\beta} + \frac{t+2}{3},
\end{equation*}
where $\bm{\sigma}_0$ and $\bm{\sigma}_f$ are the free-functions computed at $t=1$ and $t=2$, respectively.\\
The results for Problem 1 are presented in figures \ref{fig:ode3mc} and \ref{fig:ode3}, and tables \ref{tab:ode3err}. The results show that the average absolute errors for both training and testing are on the order of $\mathcal{O}(10^{-16})$. The computational time is on the order of $\mathcal{O}(10^{-3})$ seconds. 
Note that, this problem requires an iterative-least squares procedure to compute the solution \cite{NDE}. However, the computational time for each iteration is on the order of $\mathcal{O}(10^{-4})$ seconds.
\begin{table}[hp]
\begin{center}
\begin{tabular}{cccc} 
\toprule
\makecell{Test \\ Points} & \makecell{X-TFC} & \makecell{TFC}   \\ \bottomrule \midrule
{ 1 } & { 0 } & { 0 }   \\
{ 1.1 } & { 0 } & { 0 }   \\
{ 1.2 } & { 2.2$\times 10^{-16}$ } & { 0 }   \\
{ 1.3 } & { 2.2$\times 10^{-16}$ } & { 2.2$\times 10^{-16}$ }   \\
{ 1.4 } & { 2.2$\times 10^{-16}$ } & { 0 }   \\
{ 1.5 } & { 2.2$\times 10^{-16}$ } & { 2.2$\times 10^{-16}$ }   \\
{ 1.6 } & { 0 } & { 2.2$\times 10^{-16}$ }   \\
{ 1.7 } & { 2.2$\times 10^{-16}$ } & { 2.2$\times 10^{-16}$ }   \\
{ 1.8 } & { 2.2$\times 10^{-16}$ } & { 2.2$\times 10^{-16}$ }   \\
{ 1.9 } & { 0 } & { 0 }   \\
{ 2 } & { 0 } & { 0 }   \\
\bottomrule
\end{tabular}
\end{center}
\caption{Problem \#1: X-TFC and TFC absolute errors with respect the exact solution. The absolute errors of LeNN method \cite{yang} are not reported. As reported in \cite{yang}, the average absolute error of LeNN method is of $\mathcal{O}(10^{-3})$}
\label{tab:ode3err}
\end{table}

\subsubsection{Problem 2}
\begin{figure}[hp]
    \centering
    \includegraphics[width=\linewidth]{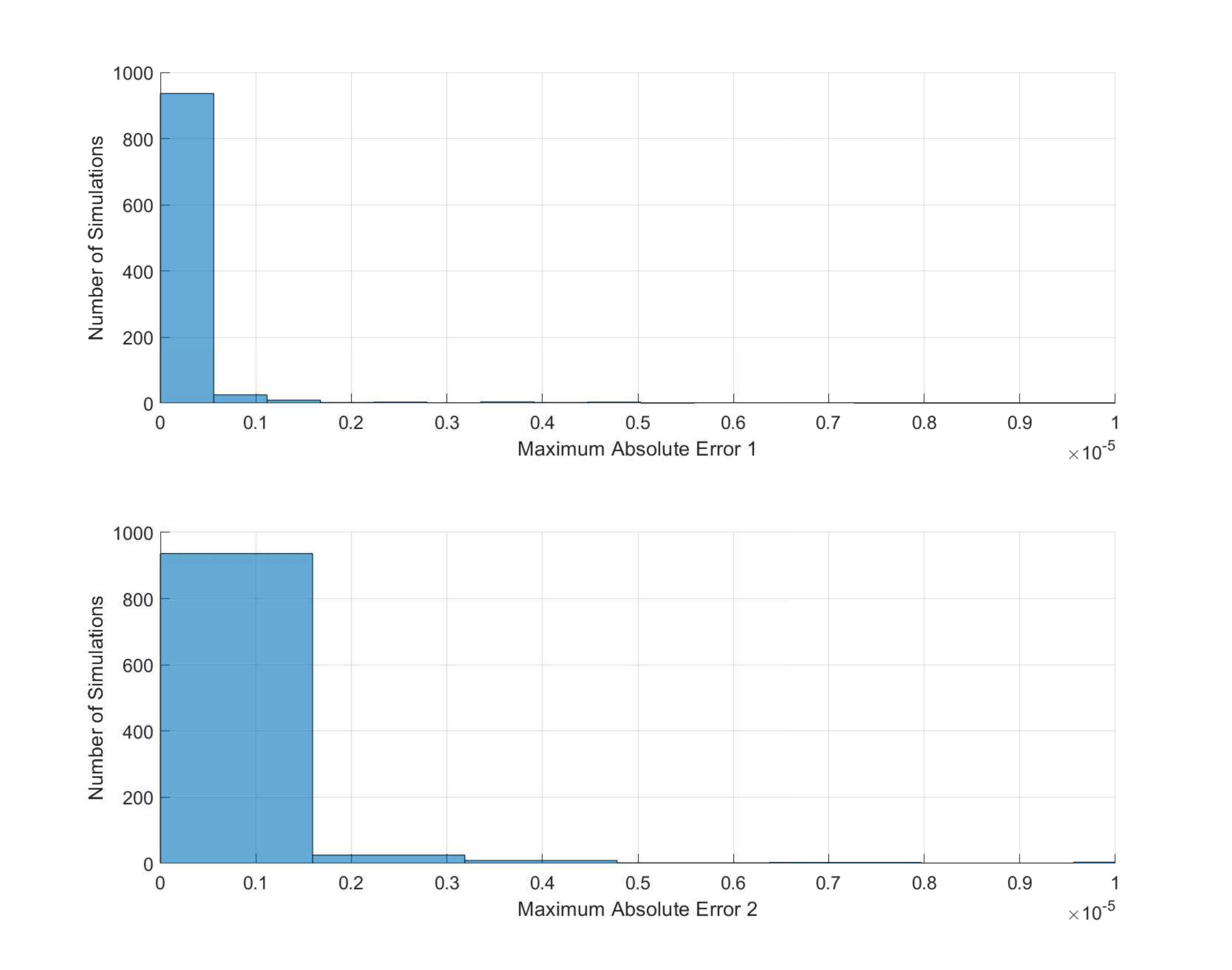}
    \caption{ histogram of $10^3$ Monte Carlo simulations for problem 2 with a \textit{logistic} activation function, $n=100$, and $L=100$.}
    \label{fig:ode4mc}
\end{figure}

\begin{figure}[hp]
    \centering
    \includegraphics[width=\linewidth]{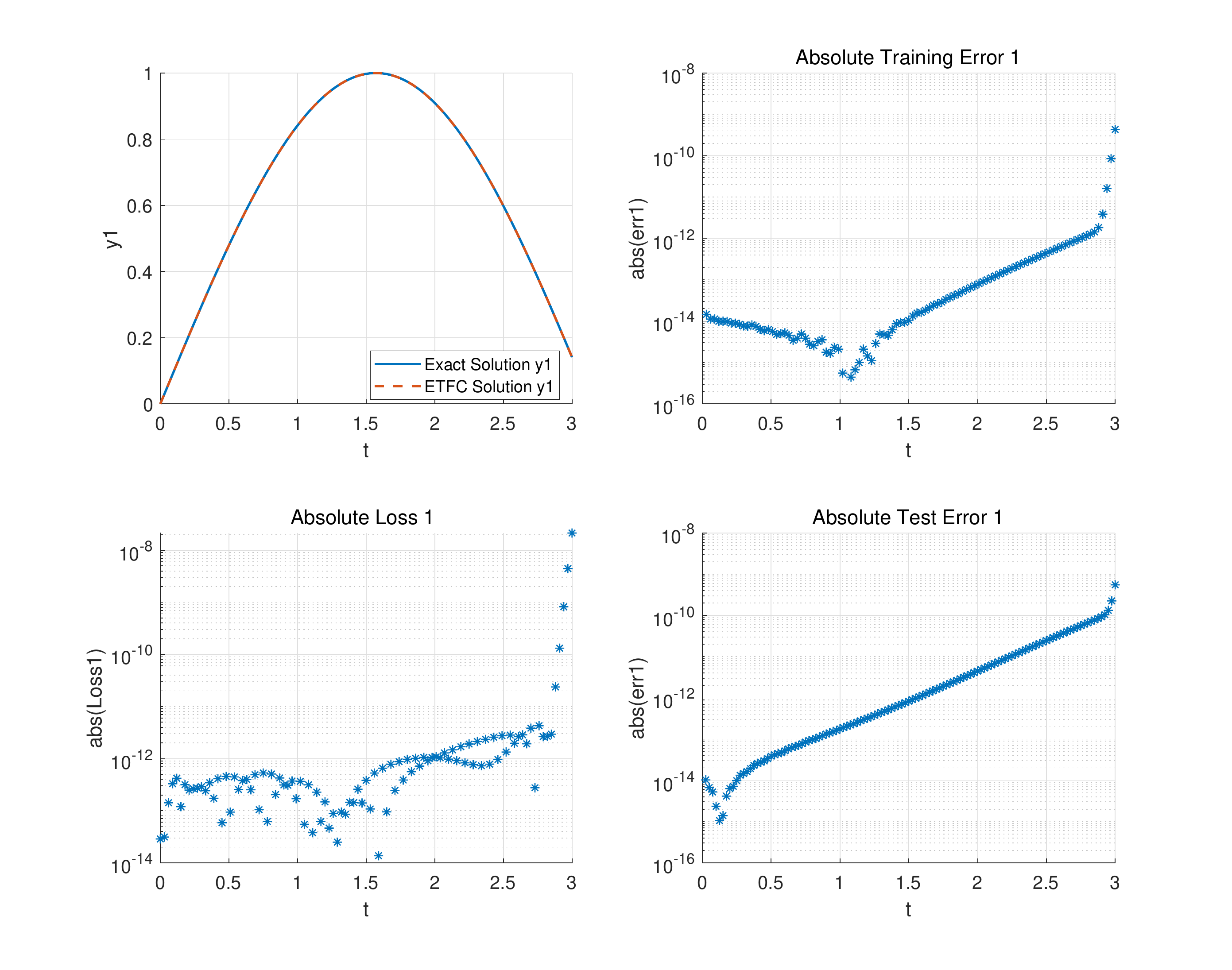}
    \caption{Solution and performances for problem 2, $y_1$: exact solution versus X-TFC solution, absolute training error, absolute loss, and absolute test error, with a \textit{logistic} activation function, $n=100$, and $L=100$. The convergence is achieved in $9$ iterations with a tolerance set to $ 10\times10^{-6} $. The computational time is $ 25.8953 $ milliseconds, the maximum absolute training error is $ 4.31822 \times10^{-10} $, the maximum absolute training loss is $ 2.13878\times10^{-8} $, and the maximum absolute test error is $ 5.5434\times10^{-10} $}
    \label{fig:ode41}
\end{figure}

\begin{figure}[hp]
    \centering
    \includegraphics[width=\linewidth]{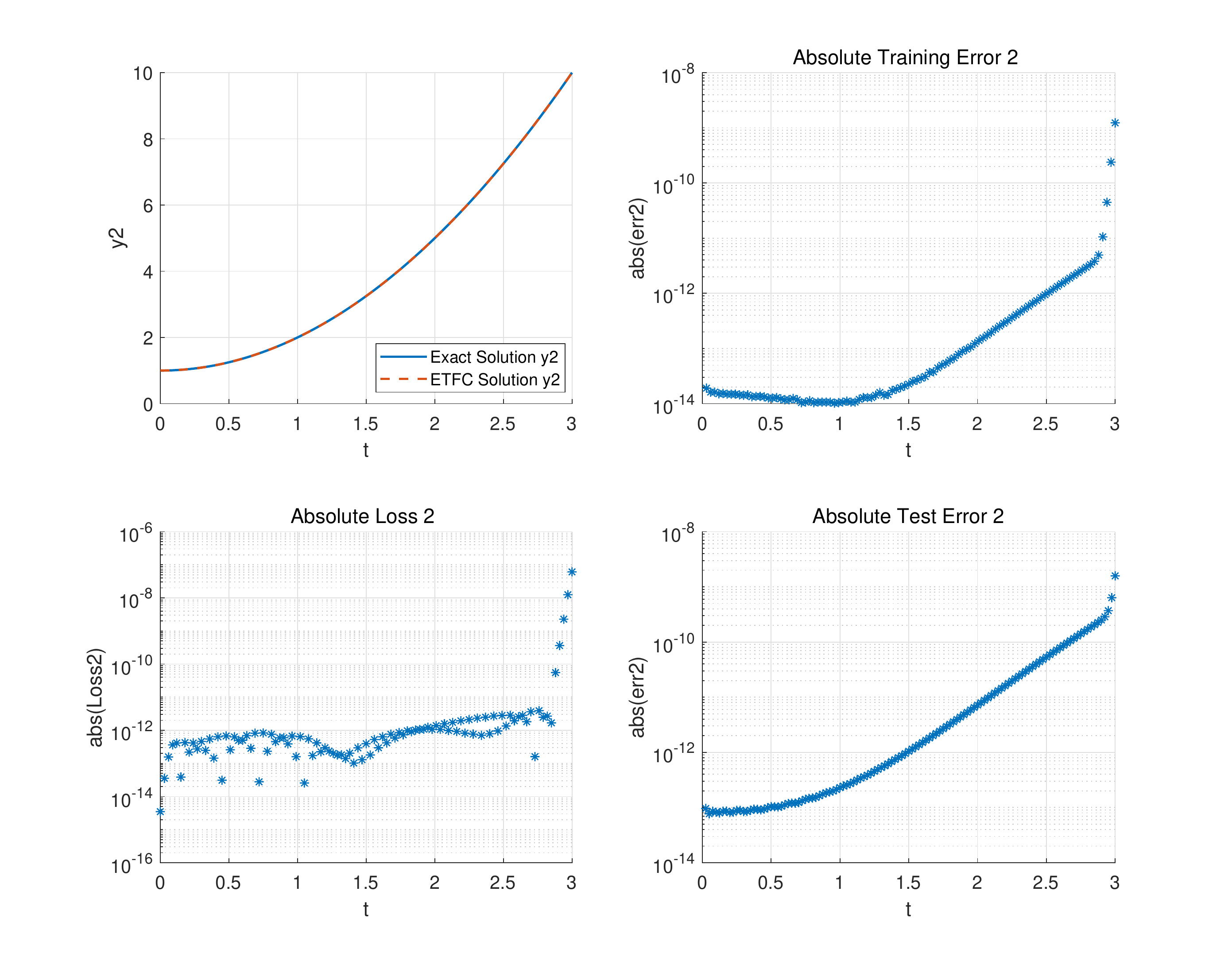}
    \caption{Solution and performances for problem 2, $y_1$: exact solution versus X-TFC solution, absolute training error, absolute loss, and absolute test error, with a \textit{logistic} activation function, $n=100$, and $L=100$. The convergence is achieved in $9$ iterations with a tolerance set to $ 10 \times10^{-6} $. The computational time is $ 25.8953 $ milliseconds, the maximum absolute training error is $ 1.23056 \times10^{-9} $, the maximum absolute training loss is $ 6.09654\times10^{-8} $, and the maximum absolute test error is $ 1.58122\times10^{-9} $}
    \label{fig:ode42}
\end{figure}

Problem 2 is a system of non-linear ODEs taken from Lagaris et al. \cite{OrigOdePde},
\begin{eqnarray*}
y_{1_t} &=& \cos t  + y^2_1 + y_2 - (1 + t^2 + \sin^2 t)  \\ 
y_{2_t} &=& 2t - (1 + t^2) \sin t + y_1 y_2
\end{eqnarray*}
subject to $y_1(0) = 0$ and $y_2(0) = 1$ where $t \in [0,3]$. The exact solutions are $ y_1(t) = \sin(t) $ and $ y_2(t) = 1 + t^2$. The constrained expressions for this problem are,
\begin{eqnarray*}
y_1 (t) &=& \left[ \bm{\sigma} - \bm{\sigma}_0 \right]\T \bm{\beta}_1 \\ 
y_2 (t) &=& \left[ \bm{\sigma} - \bm{\sigma}_0\right]\T \bm{\beta}_2 + 1
\end{eqnarray*}
where $\bm{\sigma}_0$ is the free-function computed at $t=0$.\\
The results are reported in figures \ref{fig:ode4mc}-\ref{fig:ode42}, and tables \ref{tab:ode4y1err} and \ref{tab:ode4y2err}. For $y_1$, the average absolute errors for both training and testing are on the order of $\mathcal{O}(10^{-12})$, while for $y_2$, the average absolute errors for both training and testing are on the order of $\mathcal{O}(10^{-11})$. The computational time for both $y_1$ and $y_2$ is on the order of $\mathcal{O}(10^{-2})$ seconds. 
\begin{table}[hp]
\begin{center}
\begin{tabular}{cccc} 
\toprule
\makecell{Test \\ Points} & \makecell{X-TFC} & \makecell{TFC}   \\ \bottomrule \midrule
{ 0 } & { 0 } & { 0 }   \\
{ 0.3 } & { 1.1$\times 10^{-13}$ } & { 1.2$\times 10^{-13}$  }   \\
{ 0.6 } & { 1.7$\times 10^{-13}$ } & { 2.9$\times 10^{-13}$  }   \\
{ 0.9 } & { 3.0$\times 10^{-13}$ } & { 7.2$\times 10^{-13}$  }   \\
{ 1.2 } & { 6.7$\times 10^{-13}$ } & { 1.6$\times 10^{-12}$  }   \\
{ 1.5 } & { 1.7$\times 10^{-12}$ } & { 4.1$\times 10^{-12}$  }   \\
{ 1.8 } & { 4.4$\times 10^{-12}$ } & { 1.1$\times 10^{-11}$  }   \\
{ 2.1 } & { 1.2$\times 10^{-11}$ } & { 3.0$\times 10^{-11}$  }   \\
{ 2.4 } & { 3.4$\times 10^{-11}$ } & { 8.0$\times 10^{-11}$   }   \\
{ 2.7 } & { 9.4$\times 10^{-11}$ } & { 1.8$\times 10^{-10}$  }   \\
{ 3 } & { 1.8$\times 10^{-10}$ } & { 1.6$\times 10^{-10}$  }   \\
\bottomrule
\end{tabular}
\end{center}
\caption{Problem \#2: X-TFC and TFC absolute errors on $y_1$ with respect the analytical solution. The maximum absolute error obtained with the ANN method \cite{OrigOdePde} is of $\mathcal{O}(10^{-4})$}
\label{tab:ode4y1err}
\end{table}
\begin{table}[hp]
\begin{center}
\begin{tabular}{cccc} 
\toprule
\makecell{Test \\ Points} & \makecell{X-TFC} & \makecell{TFC}   \\ \bottomrule \midrule
{ 0 } & { 0 } & { 0 }   \\
{ 0.3 } & { 6.5$\times 10^{-14}$ } & { 2.2$\times 10^{-13}$ }   \\
{ 0.6 } & { 1.3$\times 10^{-13}$ } & { 2.0$\times 10^{-13}$  }   \\
{ 0.9 } & { 2.5$\times 10^{-13}$ } & {1.2$\times 10^{-12}$   }   \\
{ 1.2 } & { 6.6$\times 10^{-13}$ } & { 1.4$\times 10^{-12}$  }   \\
{ 1.5 } & { 2.0$\times 10^{-12}$ } & { 5.5$\times 10^{-12}$ }   \\
{ 1.8 } & { 6.3$\times 10^{-12}$ } & { 1.5$\times 10^{-11}$  }   \\
{ 2.1 } & { 2.1$\times 10^{-11}$ } & { 5.2$\times 10^{-11}$ }   \\
{ 2.4 } & { 7.0$\times 10^{-11}$ } & { 1.6$\times 10^{-10}$  }   \\
{ 2.7 } & { 2.3$\times 10^{-10}$ } & { 4.5$\times 10^{-10}$  }   \\
{ 3 } & { 5.2$\times 10^{-10}$ } & { 4.7$\times 10^{-10}$ }   \\
\bottomrule
\end{tabular}
\end{center}
\caption{Problem \#4: X-TFC and TFC absolute errors on  scattered flux $y_2$ with respect the exact solution. The maximum absolute error obtained with the ANN method \cite{OrigOdePde} is of $\mathcal{O}(10^{-5})$}
\label{tab:ode4y2err}
\end{table}
\subsection{PDEs}
\begin{figure}[hp]
    \centering
    \includegraphics[width=\linewidth]{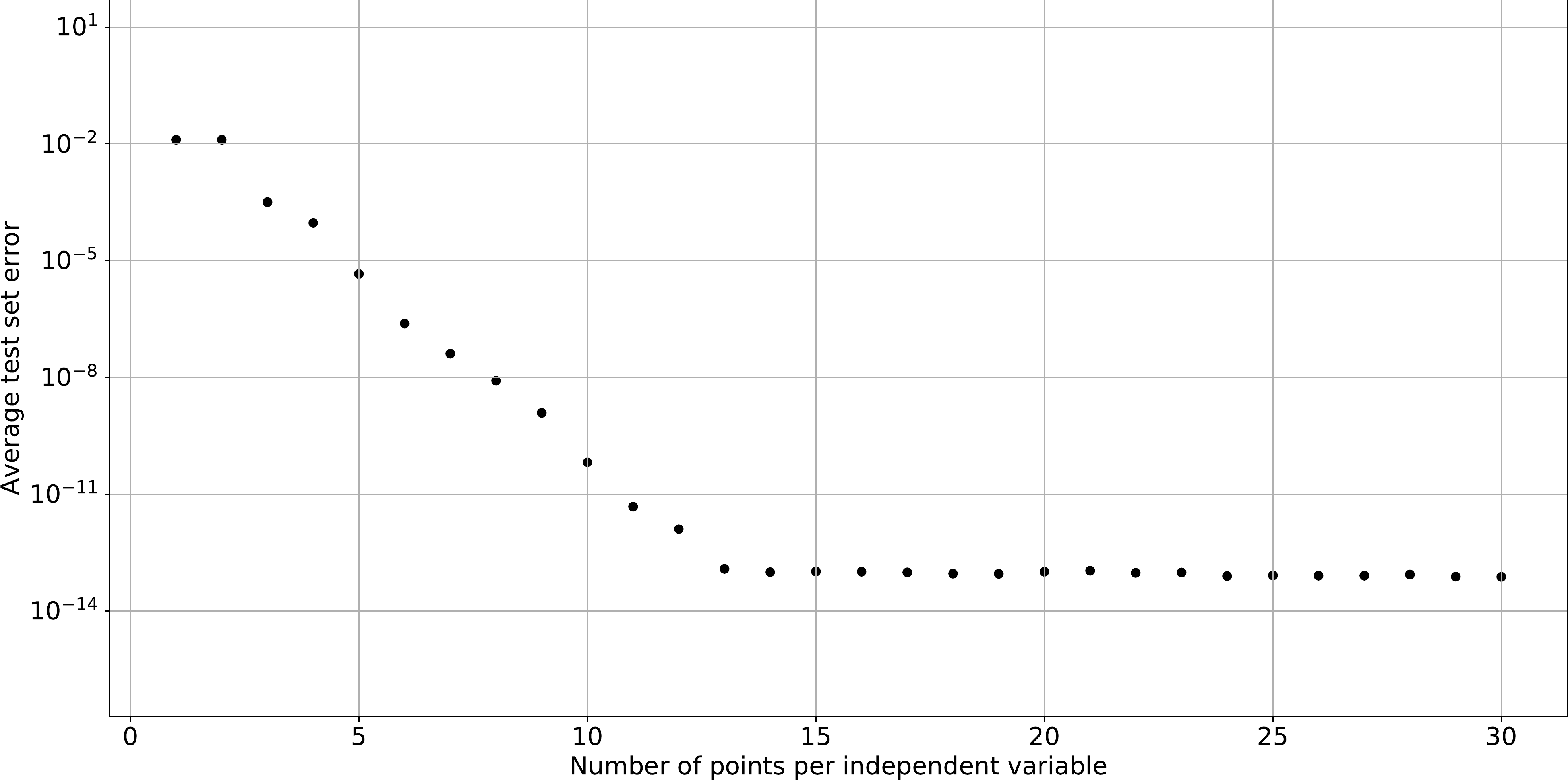}
    \caption{Average test set error as a function of the number of points per side in the grid}
    \label{fig:p1AnalysisPoints}
\end{figure}

\begin{figure}[hp]
    \centering
    \includegraphics[width=\linewidth]{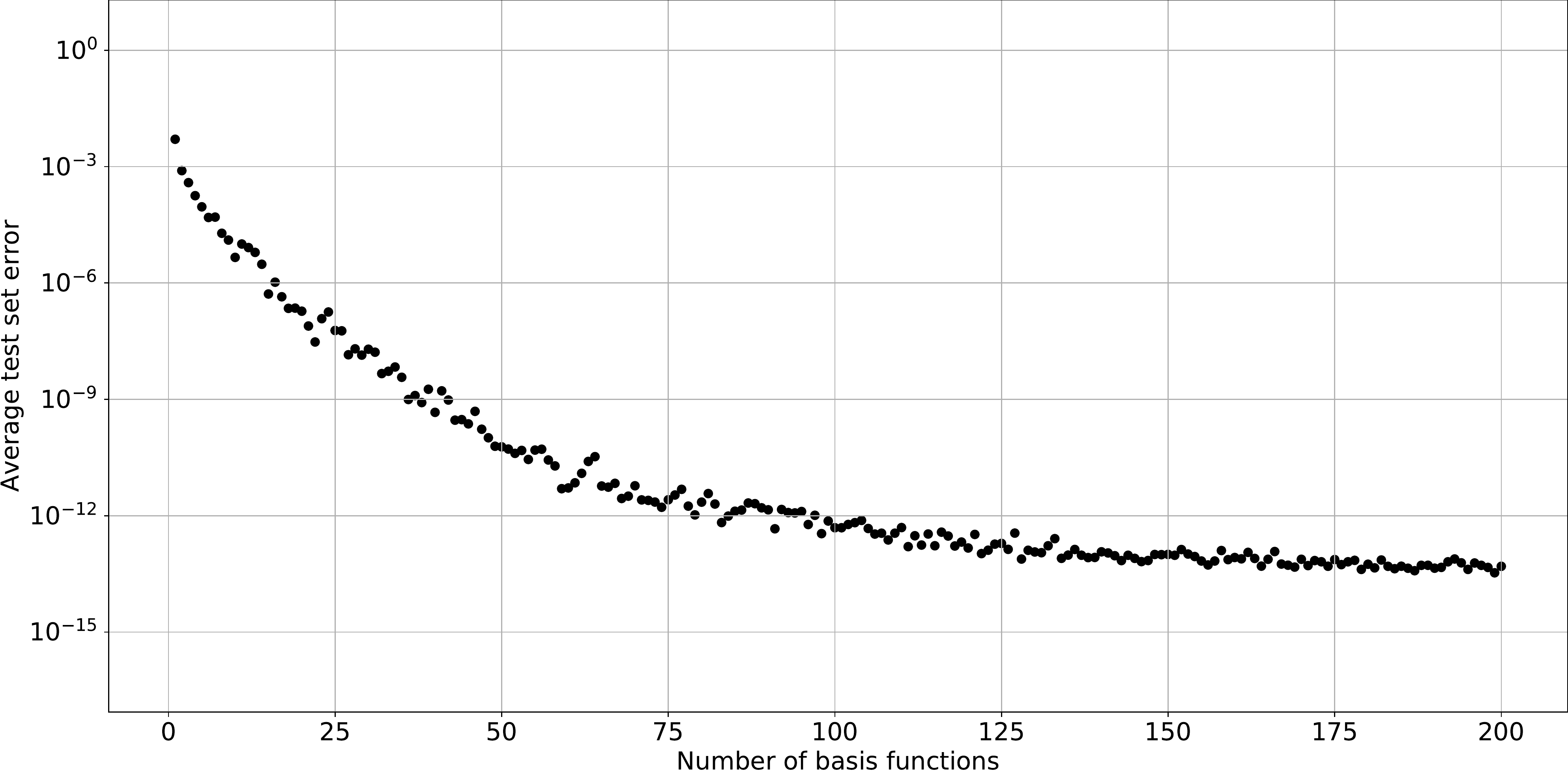}
    \caption{Average test set error as a function of the number of basis functions}
    \label{fig:p1AnalysisBasis}
\end{figure}

This section presents the results of the proposed method when applied a variety of PDEs. For each problem, the PDE and its constraints are summarized along with the relevant equations needed to construct the constrained expression. Each problem follows the same format. First, the $\mathcal{M}$ tensor and $\B{v}$ vectors are presented followed by an expanded form of the constrained expression. Problem 1 provides a reminder of the notation used in the constrained expressions, while the latter examples simply state the terms used. For a more detailed explanation on how to build constrained expression, consult Refs. \cite{TFC,M-TFC}.\\
All PDE problems shown in this article were implemented in Python and utilized the autograd package \cite{autograd}, which uses automatic differentiation \cite{autodiff} to compute the derivatives. Consequently, two specific computation times are provided, 1) the full run-time of the problem and 2) the computation time associated with the least-squares. As observed in the results, the full run-time is drastically affected by the computation overhead from autograd with full run times on the order of 1 - 100 seconds. However, the computation time for the least squares and nonlinear least squares is on the order of 1 - 100 milliseconds. Therefore, the second time reported, the one associated with the least squares, is the expected computation time if the partial derivative of the problem were computed analytically and explicitly programmed and/or if the problems were solved in a compiled language. All input weights and biases for all problems were randomly sampled from a uniform distribution, $(\B{w}_j,\,b_j)\,\sim \text{unif}(-1,1)$ where $j=1,...,L$.\\
For the PDEs, the same sensitivity analysis was performed as for the ODEs. The results of one of these studies is shown in Figs. \ref{fig:p1AnalysisPoints}  and \ref{fig:p1AnalysisBasis} for the PDE of problem number 1. Figure \ref{fig:p1AnalysisPoints} shows the solution error for the PDE of problem number 1 as a function of the number of training points in the grid. The $x$-axis in this figure shows the number of training points used per independent variable: the total number of training points used was the square of the values on the $x$-axis. Each data point in Fig. \ref{fig:p1AnalysisPoints} used 170 basis functions. Figure \ref{fig:p1AnalysisBasis} shows the solution error for the PDE of problem number 1 as a function of the number of basis functions. The number of training points used for each point in Fig. \ref{fig:p1AnalysisBasis} was 900, a $30\times 30$ grid. The results are consistent with the results obtained for the ODEs: the solution error asymptotically decreases as the number of basis functions increases, and the solution error asymptotically decreases as the the number of training points increases. 
\\
The tests in this section were performed on a MacBook Pro (2016) macOS Version 10.15.3 with a 3.3 GHz Dual-Core Intel\textsuperscript{\textregistered} Core\texttrademark \, i7 and with 16 GB of RAM. All run times were calculated using the default\_timer function in the Python \verb"timeit" package.
\subsubsection{Problem 1}
\begin{figure}[hp]
    \centering
    \includegraphics[width=.8\linewidth]{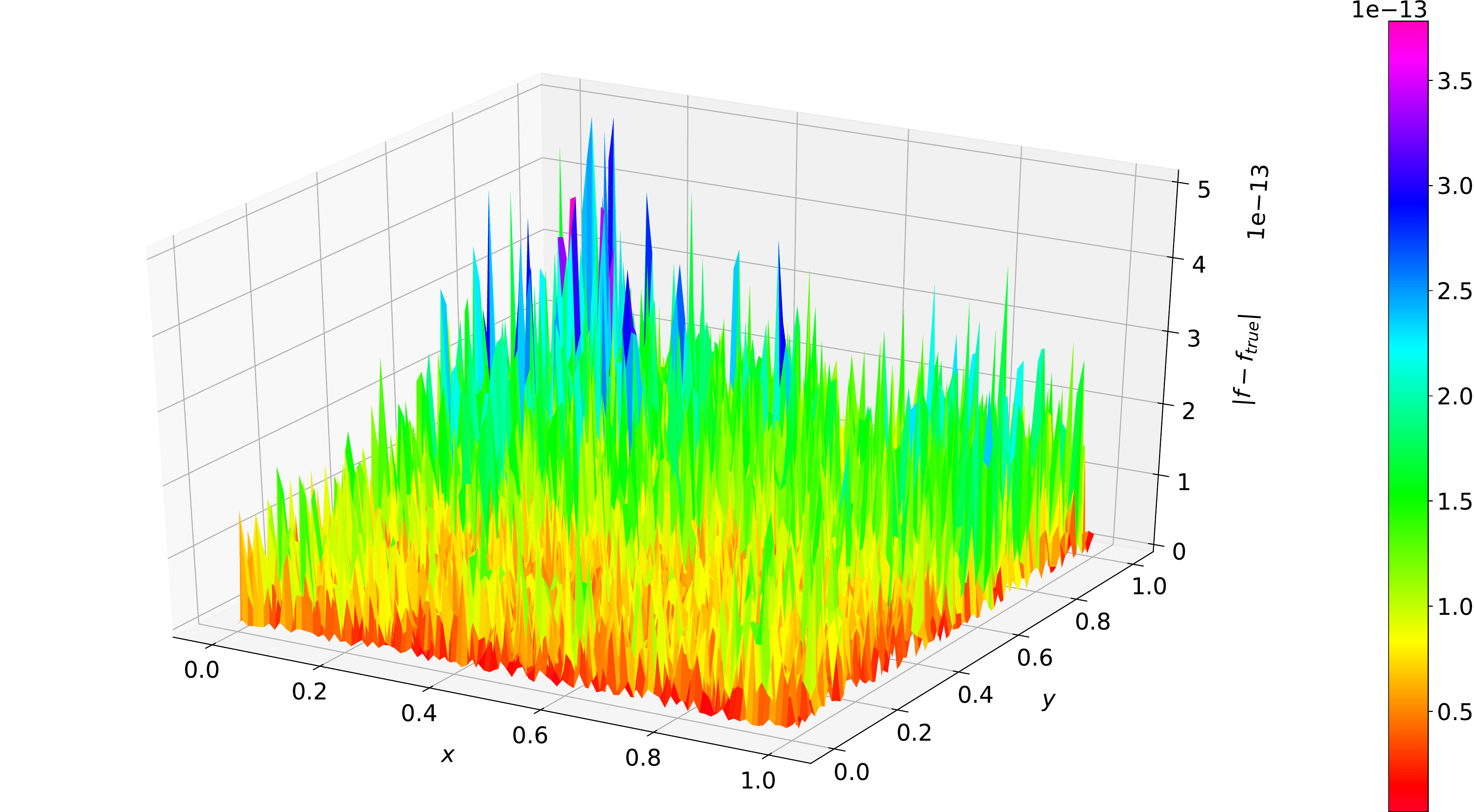}
    \caption{Solution error using X-TFC in problem 1}
    \label{fig:pde_p1}
\end{figure}

Consider the PDE solved in Largaris et al. \cite{OrigOdePde}, Mall \& Chakraverty \cite{CNN}, and Sun et al. \cite{BNN},
\begin{equation*}
    f_{xx} (x,y)+f_{yy} (x,y) = e^{-x}(x - 2 + y^3 + 6y)
\end{equation*}
where $x,y \in [0,1]$ and subject to,
\begin{eqnarray*}
    f(0,y) &=& y^3\\
    f(1,y) &=& (1+y^3)e^{-1}\\
    f(x,0) &=& xe^{-x}\\
    f(x,1) &=& e^{-x}(x+1),
\end{eqnarray*}
which has the true solution $f(x,y) = e^{-x}(x + y^3)$. Using the TFC \cite{M-TFC}, the constrained expression for the specified boundary conditions can be written in its compact form,
\begin{equation*}
    f(x,y) = g(x,y)+\B{v}(x)\T \mathcal{M}(c(x,y)-g(x,y))\B{v}(y),
\end{equation*}
where $g(x,y)$ will be defined as an ELM, the $c$ terms are defined by the constraints (e.g. $c(0,y) := f(0,y)$), and the values $c(0,0), c(0,1), c(1,0),$ and $c(1,1)$ coincide with the intersection of the constraints and are therefore defined by the constraints. Furthermore, for this problem,
\begin{equation*}
    \mathcal{M}(c(x,y)) = \begin{bmatrix} 0 & c(x,0) & c(x,1) \\ c(0,y) & -c(0,0) & -c(0,1) \\ c(1,y) & -c(1,0) & -c(1,1)\end{bmatrix}.
\end{equation*}
and
\begin{equation*}
    \B{v}(x) = \begin{Bmatrix} 1, & 1-x, & x\end{Bmatrix}\T, \quad
    \B{v}(y) = \begin{Bmatrix} 1, & 1-y, & y\end{Bmatrix}\T.
\end{equation*}
It follows that the expanded constrained expression is,
\begin{align*}
    f(x,y) =\ &g(x,y)-(x-1) \left(y (-g(0,0)+g(0,1)-1)+g(0,0)+y^3\right)+(x-1) g(0,y)\\
    &+x (y g(1,1)-(y-1) g(1,0))-x g(1,y)+(y-1) g(x,0)-y g(x,1)+\frac{x y \left(y^2-1\right)}{e}+e^{-x} (x+y)
\end{align*}
For this problem, the free-function, $g(x,y)$, was chosen to be an ELM with 170 neurons that used \verb"tanh" as the non-linear activation function. Then, the constrained expression and its derivatives were substituted into the differential equation, which converts the differential equation into an algebraic equation. To solve this algebraic equation, the problem was discretized over 30$\times$30 training points that spanned the domain. This system of equations was solved using NumPy's \verb"lstsq" function.\\
The total execution time was 3.48 seconds, and the least-squares took 8.07 milliseconds. Additionally, the training set maximum error was $3.808\times10^{-13}$, and the training set average error was $6.475\times10^{-14}$. The test set maximum error was $5.054\times10^{-13}$, and the test set average error was $7.640\times10^{-14}$. 
Figure \ref{fig:pde_p1} shows a plot of the error over the domain, and 
Table \ref{tab:p1} compares the X-TFC solution with the FEM and Refs. \cite{OrigOdePde,CNN,BNN}. 
\begin{table}[hp]
\begin{center}
\begin{tabular}{ccc} 
\toprule
\makecell{Method} & \makecell{Training Set\\Maximum Error} & \makecell{Test Set\\Maximum Error}\\ \bottomrule \midrule
{X-TFC} & {$3.8 \times 10^{-13}$} & {$5.1 \times 10^{-13}$} \\
{FEM} & {$2 \times 10^{-8}$} & {$1.5 \times 10^{-5}$} \\
{Ref. \cite{OrigOdePde}} & {$5 \times 10^{-7}$} & {$5 \times 10^{-7}$} \\
{Ref. \cite{CNN}} & {$-$} & {$3.2 \times 10^{-2}$} \\
{Ref. \cite{BNN}} & {$-$} & {$2.4 \times 10^{-4}$} \\
\bottomrule
\end{tabular}
\end{center}
\caption{Comparison of maximum training and test error of X-TFC with current state-of-the-art techniques for Problem 1.}
\label{tab:p1}
\end{table}
Figure \ref{fig:pde_p1} shows that the error is distributed approximately evenly throughout the domain. 
Table \ref{tab:p1} shows that the X-TFC method outperforms the other methods in terms of solution error by 5 to 11 orders of magnitude. 
\subsubsection{Problem 2}
\begin{figure}[hp]
    \centering
    \includegraphics[width=.75\linewidth]{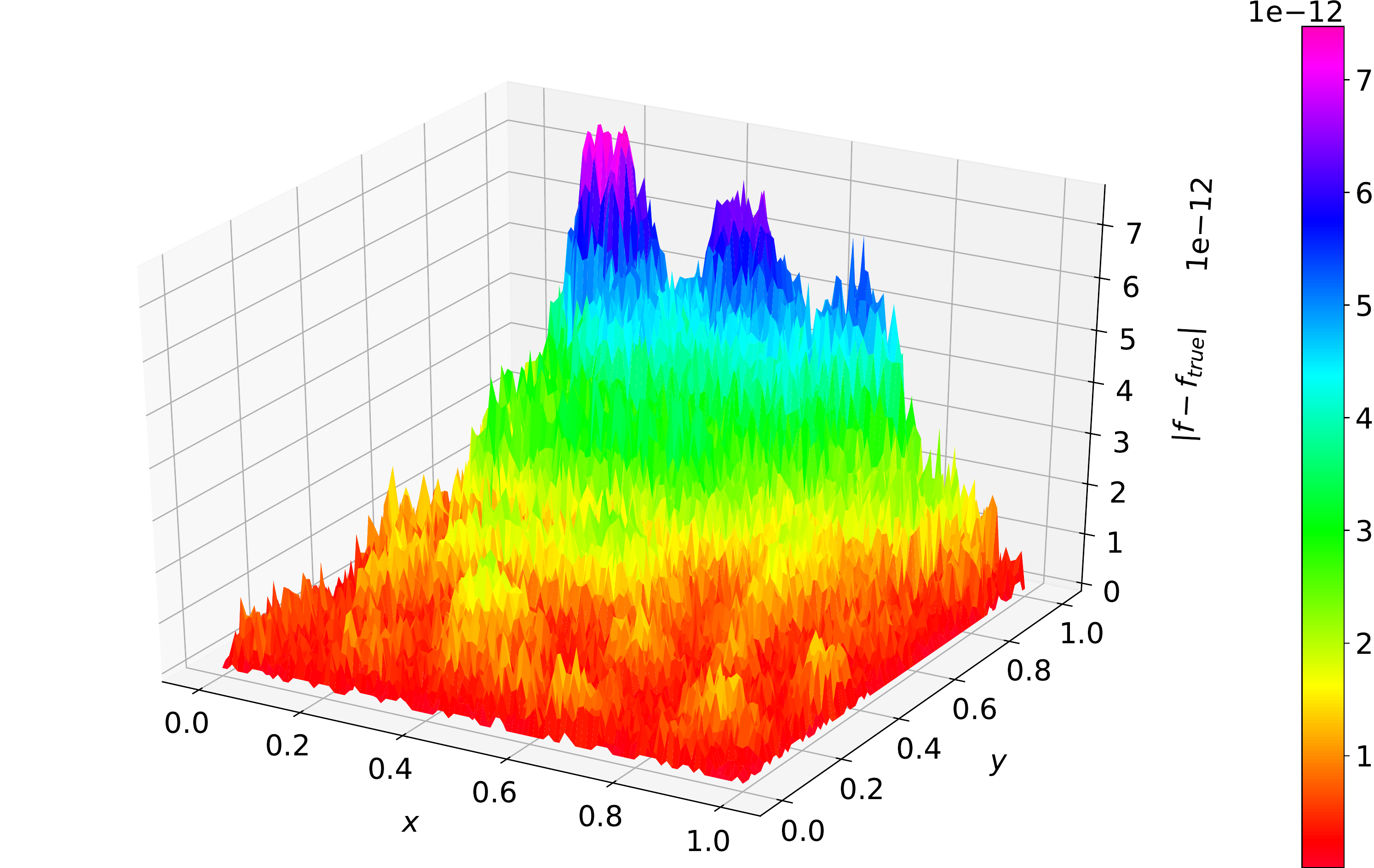}
    \caption{Solution error using X-TFC in problem 2}
    \label{fig:pde_p2}
\end{figure}

Consider the PDE solved in Largaris et al. \cite{OrigOdePde} and Mall \& Chakraverty \cite{CNN},
\begin{equation*}
    f_{xx} (x,y)+f_{yy} (x,y) = (2 - \pi^2 y ^2)\sin(\pi x) 
\end{equation*}
where $x,y \in [0,1]$ and subject to,
\begin{eqnarray*}
    f(0,y) &=& 0\\
    f(1,y) &=& 0\\
    f(x,0) &=& 0\\
    f_y(x,1) &=& 2\sin(\pi x),
\end{eqnarray*}
which has the true solution $f(x,y) = y^2 \sin(\pi x)$.
\begin{tcolorbox}[breakable,colback=white,colframe=darkgray,width=\dimexpr\textwidth\relax]
\noindent Constrained expression (compact):
\begin{equation*}
    f(x,y) = g(x,y)+\B{v}(x)\T \mathcal{M}(c(x,y)-g(x,y))\B{v}(y)
\end{equation*}
where
\begin{equation*}
    \mathcal{M}(c(x,y)) = \begin{bmatrix} 0 & c(x,0) & c_y(x,1) \\ c(0,y) & -c(0,0) & -c_y(0,1) \\ c(1,y) & -c(1,0) & -c_y(1,1)\end{bmatrix}
\end{equation*}
and
\begin{equation*}
    \B{v}(x) = \begin{Bmatrix} 1, & 1-x, & x\end{Bmatrix}\T, \quad
    \B{v}(y) = \begin{Bmatrix} 1, & 1, & y\end{Bmatrix}\T.
\end{equation*}

\noindent Constrained expression (expanded):
\begin{align*}
    f(x,y) =\ &y \left((1-x) g_y(0,1)+x g_y(1,1)-g_y(x,1)+2 \sin (\pi  x)\right)\\
    &-(1-x) g(0,y)-x g(1,y)+g(x,y)+(1-x) g(0,0)+x g(1,0)-g(x,0)
\end{align*}
\end{tcolorbox}
For this problem, the free-function was chosen to be an with 170 neurons that used \verb"tanh" as the activation function. The problem was discretized over 30$\times$30 training points that spanned the domain, and the least-squares problem was solved using NumPy's \verb"lstsq" function.\\
The total execution time was 3.54 seconds, and the least-squares took 9.34 milliseconds. Furthermore, the training set maximum error was $6.332\times10^{-12}$, and the training set average error was $1.187\times10^{-12}$. The test set maximum error was $7.581\times10^{-12}$, and the test set average error was $1.322\times10^{-12}$. 
Figure \ref{fig:pde_p2} shows a plot of the error over the domain, and 
Table \ref{tab:p2} compares the X-TFC solution with the FEM and Refs. \cite{OrigOdePde} and \cite{CNN}. 
\begin{table}[hp]
\begin{center}
\begin{tabular}{ccc} 
\toprule
\makecell{Method} & \makecell{Training Set\\Maximum Error} & \makecell{Test Set\\Maximum Error}\\ \bottomrule \midrule
{X-TFC} & {$6.3 \times 10^{-12}$} & {$7.6 \times 10^{-12}$} \\
{FEM} & {$7 \times 10^{-7}$} & {$4 \times 10^{-5}$} \\
{Ref. \cite{OrigOdePde}} & {$6 \times 10^{-6}$} & {$6 \times 10^{-6}$} \\
{Ref. \cite{CNN}} & {$-$} & {$3 \times 10^{-3}$} \\
\bottomrule
\end{tabular}
\end{center}
\caption{Comparison of maximum training and test error of X-TFC with current state-of-the-art techniques for Problem 2.}
\label{tab:p2}
\end{table}
Figure \ref{fig:pde_p2} shows that the solution error is lower near where constraints are defined on the boundary value than where they are defined on the derivative. Intuitively, this makes sense as the TFC constrained expressions guarantees there will be no error in the solution value for constraints defined on the boundary value, but not for constraints on the derivative (there it only garuantees that the deriative will have no error). 
Table \ref{tab:p2} shows that the X-TFC method outperforms all other methods in terms of accuracy by 5 to 9 orders of magnitude.
\subsubsection{Problem 3}
\begin{figure}[hp]
    \centering
    \includegraphics[width=.75\linewidth]{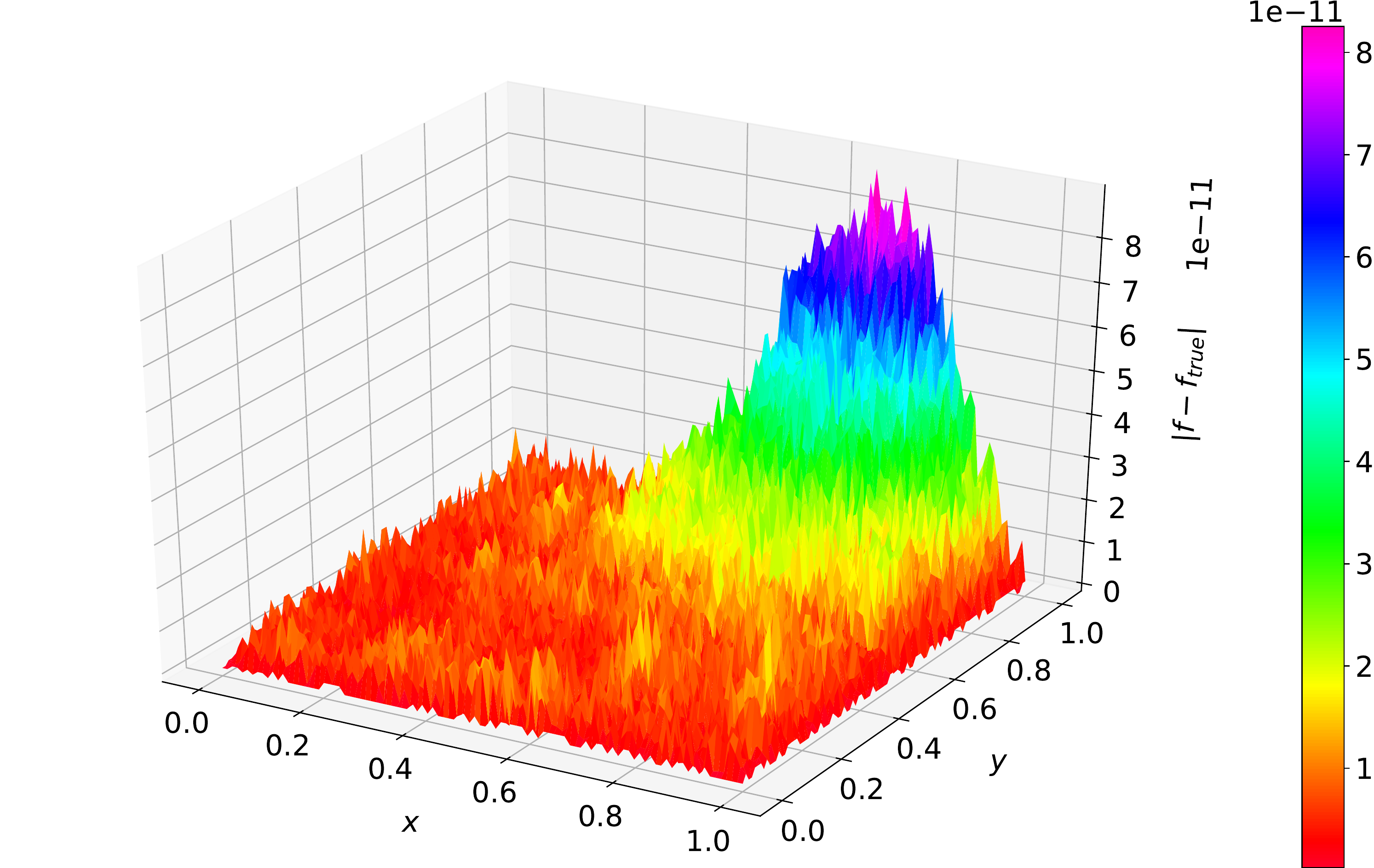}
    \caption{Solution error using X-TFC in problem 3}
    \label{fig:pde_p3}
\end{figure}

Consider the PDE solved in Largaris et al. \cite{OrigOdePde},
\begin{equation*}
   f_{xx} (x,y)+f_{yy} (x,y) + f(x,y) f_y(x,y) = \sin(\pi x) \Big(2 - \pi^2y^2 + 2y^3 \sin(\pi x)\Big)
\end{equation*}
where $x,y \in [0,1]$ and subject to,
\begin{eqnarray*}
    f(0,y) &=& 0\\
    f(1,y) &=& 0\\
    f(x,0) &=& 0\\
    f_y(x,1) &=& 2\sin(\pi x),
\end{eqnarray*}
which has the true solution $f(x,y) = y^2 \sin(\pi x)$.
\begin{tcolorbox}[breakable,colback=white,colframe=darkgray,width=\dimexpr\textwidth\relax]
\noindent Constrained expression (compact):
\begin{equation*}
    f(x,y) = g(x,y)+\B{v}(x)\T \mathcal{M}(c(x,y)-g(x,y))\B{v}(y)
\end{equation*}
where
\begin{equation*}
    \mathcal{M}(c(x,y)) = \begin{bmatrix} 0 & c(x,0) & c_y(x,1) \\ c(0,y) & -c(0,0) & -c_y(0,1) \\ c(1,y) & -c(1,0) & -c_y(1,1)\end{bmatrix}
\end{equation*}
and
\begin{equation*}
    \B{v}(x) = \begin{Bmatrix} 1, & 1-x, & x\end{Bmatrix}\T, \quad
    \B{v}(y) = \begin{Bmatrix} 1, & 1, & y\end{Bmatrix}\T.
\end{equation*}

\noindent Constrained expression (expanded):
\begin{align*}
    f(x,y) =\ &y \left((1-x) g_y(0,1)+x g_y(1,1)-g_y(x,1)+2 \sin (\pi  x)\right)\\
    &-(1-x) g(0,y)-x g(1,y)+g(x,y)+(1-x) g(0,0)+x g(1,0)-g(x,0)
\end{align*}
\end{tcolorbox}
For this problem, the free-function was chosen to be an ELM with 150 neurons that used \verb"tanh" as the activation function. The problem was discretized over 20$\times$20 training points that spanned the domain, and each iteration of the non-linear least-squares was solved using NumPy's \verb"lstsq" function.
The total execution time was 22.48 seconds, and the nonlinear least-squares, which needed 10 iterations, took 52.6 milliseconds. In addition, the training set maximum error was $7.634\times10^{-11}$, and the training set average error was $9.497\times10^{-12}$. The test set maximum error was $8.977\times10^{-11}$, and the test set average error was $1.068\times10^{-11}$. 
Figure \ref{fig:pde_p3} shows a plot of the error over the domain, and 
Table \ref{tab:p3} compares the X-TFC solution with the FEM method and Ref. \cite{OrigOdePde}.
\begin{table}[hp]
\begin{center}
\begin{tabular}{ccc} 
\toprule
\makecell{Method} & \makecell{Training Set\\Maximum Error} & \makecell{Test Set\\Maximum Error}\\ \bottomrule \midrule
{X-TFC} & {$8.8 \times 10^{-11}$} & {$9.0 \times 10^{-11}$} \\
{FEM} & {$6 \times 10^{-7}$} & {$4 \times 10^{-5}$} \\
{Ref. \cite{OrigOdePde}} & {$1.5 \times 10^{-5}$} & {$1.5 \times 10^{-5}$} \\
\bottomrule
\end{tabular}
\end{center}
\caption{Comparison of maximum training and test error of X-TFC with current state-of-the-art techniques for Problem 3.}
\label{tab:p3}
\end{table}
As in problem 2, Fig. \ref{fig:pde_p3} shows illustrates that the solution error is lower near where constraints are defined on the boundary value than where they are defined on the derivative, because the TFC constrained expressions guarantees there will be no error in the solution value for constraints defined on the boundary. 
Table \ref{tab:p3} shows that the X-TFC method outperforms all other methods in terms of accuracy by 4 to 6 orders of magnitude. 
\subsubsection{Problem 4 (1D Time-Dependent Heat Equation)}
\begin{figure}[hp]
    \centering
    \includegraphics[width=.75\linewidth]{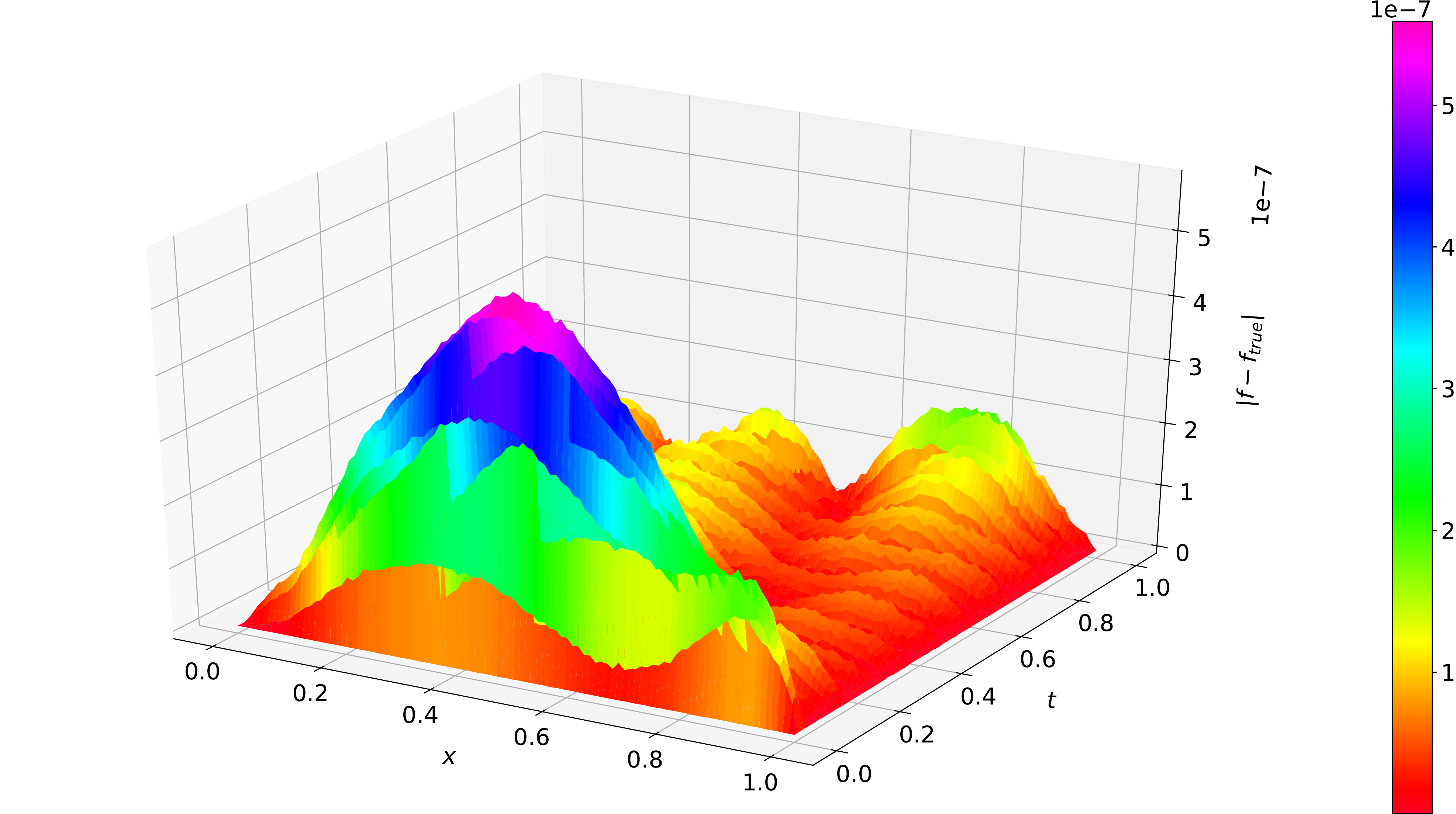}
    \caption{Solution error using X-TFC in problem 4}
    \label{fig:pde_p4}
\end{figure}

\begin{equation*}
     f_{xx} (x,t) = \kappa f_t(x,t)
\end{equation*}
where $x,t \in [0,1]\times[0,1]$ and $\kappa=1$, subject to   
\begin{eqnarray*}
    f(0,t) &=& 0\\
    f(1,t) &=& 0\\
    f(x,0) &=& \sin(\pi x),
\end{eqnarray*}
which has the true solution $f(x,t) = \sin(\pi x)e^{-\pi^2t}$. 
\begin{tcolorbox}[breakable,colback=white,colframe=darkgray,width=\dimexpr\textwidth\relax]
\noindent Constrained expression (compact):
\begin{equation*}
    f(x,t) = g(x,t)+\B{v}(x)\T \mathcal{M}\Big(c(x,t)-g(x,t)\Big)\B{v}(t)
\end{equation*}
where
\begin{equation*}
    \mathcal{M}\Big(c(x,t)\Big) = \begin{bmatrix} 0 & c(x,0)  \\ c(0,t) & -c(0,0) \\ c(L,t) & -c(L,0) \end{bmatrix}
\end{equation*}
and
\begin{equation*}
    v(x) = \begin{Bmatrix} 1, & 1-x, & x \end{Bmatrix}\T, \quad
    v(t) = \begin{Bmatrix} 1, & 1\end{Bmatrix}\T.
\end{equation*}

\noindent Constrained expression (expanded):
\begin{align*}
    f(x,t) =\ & g(x,t) + (x-1) g(0,t)-x g(1,t)-x g(0,0)+x g(1,0)-g(x,0)+g(0,0)+\sin (\pi  x)
\end{align*}
\end{tcolorbox}
For this problem, the free-function was chosen to be an ELM with 196 neurons that used \verb"tanh" as the activation function. The problem was discretized over 30$\times$30 training points that spanned the domain, and the least-squares problem was solved using NumPy's \verb"lstsq" function.
The total execution time was 3.21 seconds, and the least-squares took 9.90 milliseconds. Additionally, the training set max error was $5.611\times10^{-7}$, and the training set average error was $6.551\times10^{-8}$. The test set maximum error was $5.785\times10^{-7}$, and the test set average error was $6.928\times10^{-8}$. 
Figure \ref{fig:pde_p4} shows a plot of the error over the domain.
Figure \ref{fig:pde_p4} shows that the error is larger at smaller time steps. Of course, the solution error is zero at $t=0$, because the TFC constrained expression guarantees the constraint $c(x,0)$ is satisfied. The reason the solution error, in general, is larger for smaller values of $t$ is the solution is changing much more rapidly for smaller values of $t$ than larger values of $t$ (i.e. the gradients for smaller values of $t$ are larger than the gradients for larger values of $t$).
\subsubsection{Problem 5 (2D Time-Dependent Heat Equation)}
\begin{equation*}
     f_{xx}(x,y,t)+f_{yy}(x,y,t) = \kappa f_t(x,y,t)
\end{equation*}
where $x,y,t \in [0,L]\times[0,H]\times[0,1]$, subject to   
\begin{eqnarray*}
    f(0,y,t) &=& 0\\
    f(L,y,t) &=& 0\\
    f(x,0,t) &=& 0\\
    f(x,H,t) &=& 0\\
    f(x,y,0) &=& \sin\Big(\frac{\pi x}{L}\Big)\sin\Big(\frac{\pi y}{H}\Big),
\end{eqnarray*}
which has the true solution $f(x,y,t) = \sin\Big(\frac{\pi x}{L}\Big)\sin\Big(\frac{\pi y}{H}\Big)e^{-\Big(\frac{\pi^2}{L^2}+\frac{\pi^2}{H^2}\Big)t}$. Here, the values $L=2$, $H=1$, and $\kappa=1$ are used. 
\begin{tcolorbox}[breakable,colback=white,colframe=darkgray,width=\dimexpr\textwidth\relax]
\noindent Constrained expression (compact):
\begin{equation*}
    f(x,y,t) = g(x,y,t)+ \mathcal{M}_{ijk}\Big(c(x,y,t)-g(x,y,t)\Big)v_i(x)v_j(y)v_k(t)
\end{equation*}
where
\begin{align*}
    \mathcal{M}_{ij1}\Big(c(x,y,t)\Big) &= \begin{bmatrix} 0 & c(x,0,t) & c(x,H,t) \\ 
    c(0,y,t) & -c(0,0,t) & -c(0,H,t) \\ c(L,y,t) & -c(L,0,t) & -c(L,H,t)\end{bmatrix}\\
    \mathcal{M}_{ij2}\Big(c(x,y,t)\Big) &= \begin{bmatrix} c(x,y,0) & -c(x,0,0) & -c(x,H,0) \\ 
    -c(0,y,0) & c(0,0,0) & c(0,H,0) \\ -c(L,y,0) & c(L,0,0) & c(L,H,0)\end{bmatrix}
\end{align*}
and
\begin{equation*}
    v_i(x) = \begin{Bmatrix} 1, & \frac{L-x}{L}, & \frac{x}{L} \end{Bmatrix}\T, \quad
    v_j(y) = \begin{Bmatrix} 1, & \frac{H-y}{H}, & \frac{y}{H} \end{Bmatrix}\T, \quad
    v_k(t) = \begin{Bmatrix} 1, & 1 \end{Bmatrix}\T.
\end{equation*}

\noindent Constrained expression (expanded):
\begin{equation*}
\begin{aligned}
   f(x,y,t) &=\ \frac{(H-y) \left(\frac{(L-x) g(0,0,t)}{L}+\frac{x g(L,0,t)}{L}-g(x,0,t)\right)}{H}+\frac{y \left(\frac{(L-x) g(0,H,t)}{L}+\frac{x g(L,H,t)}{L}-g(x,H,t)\right)}{H}\\
   &+\frac{(H-y) \left(-\frac{g(0,0,0) (L-x)}{L}-\frac{x g(L,0,0)}{L}+g(x,0,0)\right)}{H}+\frac{y \left(-\frac{(L-x) g(0,H,0)}{L}-\frac{x g(L,H,0)}{L}+g(x,H,0)\right)}{H}\\
   &-\frac{(L-x) g(0,y,t)}{L}-\frac{x g(L,y,t)}{L}+\frac{(L-x) g(0,y,0)}{L}+\frac{x g(L,y,0)}{L}+g(x,y,t)-g(x,y,0)\\
   &+\sin \left(\frac{\pi  y}{H}\right) \sin \left(\frac{\pi  x}{L}\right)
\end{aligned}
\end{equation*}
\end{tcolorbox}
For this problem, the free-function was chosen to be an ELM with 400 neurons that used \verb"tanh" as the activation function. The problem was discretized over 13$\times$13$\times$13 training points that spanned the domain, and the least-squares problem was solved using NumPy's \verb"lstsq" function.\\
The total execution time was 159.3 seconds, and the nonlinear least-squares took 55.4 milliseconds. Furthermore, the training set maximum error was $3.985\times10^{-4}$, and the training set average error was $1.705\times10^{-5}$. The test set maximum error was $4.025\times10^{-4}$, and the test set average error was $2.090\times10^{-5}$.
\subsubsection{Problem 6 (Non-linear 2D time-dependent PDE)}
\begin{equation*}
    z_t(x,y,t)z_x(x,y,t)+z_y(x,y,t) = t^2+(t-1)x+2\pi x \cos(2\pi xy)+(2ty+xy)\Big((t-1)y +2\pi y\cos(2\pi xy)\Big)
\end{equation*}
where $x,y,t \in [0,1]$, subject to   
\begin{eqnarray*}
    z(0,y,t) &=& t^2y\\
    z(x,0,t) &=& 0\\
    z(x,y,1) &=& y+\sin(2\pi xy),
\end{eqnarray*}
which has the true solution $z(x,y,t) = \sin(2\pi xy)+t^2y+(t-1)xy$.
\begin{tcolorbox}[breakable,colback=white,colframe=darkgray,width=\dimexpr\textwidth\relax]
\noindent Constrained expression (compact):
\begin{equation*}
    f(x,y,t) = g(x,y,t)+ \mathcal{M}_{ijk}\Big(c(x,y,t)-g(x,y,t)\Big)v_i(x)v_j(y)v_k(t)
\end{equation*}
where
\begin{align*}
    \mathcal{M}_{ij1}\Big(c(x,y,t)\Big) &= \begin{bmatrix} 0 & c(x,0,t) \\ 
    c(0,y,t) & -c(0,0,t) \end{bmatrix}\\
    \mathcal{M}_{ij2}\Big(c(x,y,t)\Big) &= \begin{bmatrix} c(x,y,1) & -c(x,0,1) \\ 
    -c(0,y,1) & c(0,0,1)\end{bmatrix}
\end{align*}
and
\begin{equation*}
    v_i(x) = \begin{Bmatrix} 1, & 1 \end{Bmatrix}\T, \quad
    v_j(y) = \begin{Bmatrix} 1, & 1 \end{Bmatrix}\T, \quad
    v_k(t) = \begin{Bmatrix} 1, & 1 \end{Bmatrix}\T.
\end{equation*}

\noindent Constrained expression (expanded):
\begin{equation*}
\begin{aligned}
   z(x,y,t) =\ &g(x,y,t)-g(x,0,t)-g(0,y,t)+g(0,0,t)-g(x,y,1)+g(x,0,1)+g(0,y,1)-g(0,0,1)\\
   &+t^2 y+\sin (2 \pi  x y)
\end{aligned}
\end{equation*}
\end{tcolorbox}
For this problem, the free-function was chosen to be an ELM with 255 neurons that used \verb"tanh" as the activation function. The problem was discretized over 8$\times$8$\times$8 training points that spanned the domain, and each iteration of the non-linear least-squares was solved using NumPy's \verb"lstsq" function.\\
The total execution time was 32.19 seconds, and the nonlinear least-squares, which needed 10 iterations, took 0.140 seconds. In addition, the training set maximum error was $1.657\times10^{-8}$, and the training set average error was $1.006\times10^{-9}$. The test set maximum error was $1.657\times10^{-8}$, and the test set average error was $1.018\times10^{-9}$.
\subsubsection{Problem 7 (Non-linear 3D time-dependent PDE)}
\begin{equation*}
\begin{aligned}
    f_x(x,y,z,t)f_y(x,y,z,t)f_z(x,y,z,t)+f_{tt}(x,y,z,t) =\ & \left((t-1) t x (z-1)+x^2 \cos \left(x^2 y\right)+\frac{3}{2} x \sqrt{y} z\right)\cdot\\
    &\left((t-1) t y (z-1)+2 x y \cos \left(x^2 y\right)+y^{3/2} z\right)\cdot \\
    &\left(2 \pi  t^2 \cos(2 \pi  z)+(t-1) t x y+x y^{3/2}\right)+2 x y (z-1)\\
    &+2 \sin(2 \pi  z)
\end{aligned}
\end{equation*}
where $x,y,t \in [0,1]$, subject to   
\begin{eqnarray*}
    f(0,y,z,t) &=& t^2 \sin(2 \pi  z)\\
    f(x,0,z,t) &=& t^2 \sin(2 \pi  z)\\
    f(x,y,1,t) &=& \sin\left(x^2 y\right)+x \sqrt{y}\\
    f(x,y,z,0) &=& \sin\left(x^2 y\right)+x \sqrt{y} z\\
    f(x,y,z,1) &=& \sin\left(x^2 y\right)+x \sqrt{y} z+\sin(2 \pi  z),
\end{eqnarray*}
which has the true solution $f(x,y,z,t) = t^2 \sin (2 \pi  z)+\sin \left(x^2 y\right)+x y^{3/2} z + xyt(z-1)(t-1)$.

\begin{tcolorbox}[breakable,colback=white,colframe=darkgray,width=\dimexpr\textwidth\relax]
\noindent Constrained expression (compact):
\begin{equation*}
    f(x,y,z,t) = g(x,y,z,t)+ \mathcal{M}_{ijkl}\Big(c(x,y,z,t)-g(x,y,z,t)\Big)v_i(x)v_j(y)v_k(z)v_l(t)
\end{equation*}
where
\begin{align*}
    \mathcal{M}_{ij11}\Big(c(x,y,z,t)\Big) &= \begin{bmatrix} 0 & c(x,0,z,t) \\ 
    c(0,y,z,t) & -c(0,0,z,t) \end{bmatrix}\\
    \mathcal{M}_{ij12}\Big(c(x,y,z,t)\Big) &= \begin{bmatrix} c(x,y,z,0) & -c(x,0,z,0) \\ 
    -c(0,y,z,0) & c(0,0,z,0) \end{bmatrix}\\
    \mathcal{M}_{ij13}\Big(c(x,y,z,t)\Big) &= \begin{bmatrix} c(x,y,z,1) & -c(x,0,z,1) \\ 
    -c(0,y,z,1) & c(0,0,z,1) \end{bmatrix}\\
    \mathcal{M}_{ij21}\Big(c(x,y,z,t)\Big) &= \begin{bmatrix} c(x,y,1,t) & -c(x,0,1,t) \\ 
    -c(0,y,1,t) & c(0,0,1,t) \end{bmatrix}\\
    \mathcal{M}_{ij22}\Big(c(x,y,z,t)\Big) &= \begin{bmatrix} -c(x,y,1,0) & c(x,0,1,0) \\ 
    c(0,y,1,0) & -c(0,0,1,0) \end{bmatrix}\\
    \mathcal{M}_{ij23}\Big(c(x,y,z,t)\Big) &= \begin{bmatrix} -c(x,y,1,1) & c(x,0,1,1) \\ 
    c(0,y,1,1) & -c(0,0,1,1) \end{bmatrix}\\
\end{align*}
and
\begin{equation*}
    v_i(x) = \begin{Bmatrix} 1, & 1 \end{Bmatrix}\T \quad
    v_j(y) = \begin{Bmatrix} 1, & 1 \end{Bmatrix}\T \quad
    v_k(z) = \begin{Bmatrix} 1, & 1 \end{Bmatrix}\T \quad
    v_l(t) = \begin{Bmatrix} 1, & 1-t, & t \end{Bmatrix}\T.
\end{equation*}
\noindent Constrained expression (expanded):
\begin{equation*}
\begin{aligned}
   f(x,y,z,t) =\ &g(x,y,z,t)+(1-t) \Big(-g(x,y,z,0)+g(x,y,1,0)+g(x,0,z,0)-g(x,0,1,0)+g(0,y,z,0)\\
   &-g(0,y,1,0)-g(0,0,z,0)+g(0,0,1,0)+x y^{3/2} z-x y^{3/2}\Big)+t \Big(-g(x,y,z,1)\\
   &+g(x,y,1,1)+g(x,0,z,1)-g(x,0,1,1)+g(0,y,z,1)-g(0,y,1,1)-g(0,0,z,1)\\
   &+g(0,0,1,1)+x y^{3/2} z-x y^{3/2}\Big)-g(x,y,1,t)-g(x,0,z,t)+g(x,0,1,t)-g(0,y,z,t)\\
   &+g(0,y,1,t)+g(0,0,z,t)-g(0,0,1,t)+t^2 \sin (2 \pi  z)+\sin \left(x^2 y\right)+x y^{3/2}
\end{aligned}
\end{equation*}
\end{tcolorbox}
For this problem, the free-function was chosen to be an ELM with 340 neurons that used \verb"tanh" as the activation function. The problem was discretized over 5$\times$5$\times$5$\times$5 training points that spanned the domain, and each iteration of the non-linear least-squares was solved using NumPy's \verb"lstsq" function.\\
The total execution time was 321.9 seconds, and the nonlinear least-squares, which needed 10 iterations, took 0.229 seconds. Additionally, the training set maximum error was $2.744\times10^{-5}$, and the training set average error was $6.641\times10^{-7}$. The test set maximum error was $2.978\times10^{-5}$, and the test set average error was $8.082\times10^{-7}$.
\section{Conclusions}
In this article, a novel, accurate, and robust physics-informed (PI) method for solving problems involving parametric DEs called the \textit{Extreme Theory of Functional Connection}, or \textit{Extreme-TFC} (X-TFC) was developed. Although this article focuses on the solution of \textit{exact problems} (e.g. problems where the modeling error is negligible), X-TFC can also handle \textit{data-driven solutions} and \textit{data-driven discovery} of parametric DEs.
X-TFC is a synergy of the standard TFC method developed by Mortari et al. \cite{TFC,LDE,NDE}, where the latent solution of the DE is approximated by a constrained expression, which analytically satisfies the constraints while maintaining a free-function that can be used to satisfy the parametric DE, and the classic physics-informed neural network (PINN) methods proposed by Raissi et al. \cite{raissi} and \cite{ModernPDE}, where the free-function is chosen as a NN. In X-TFC, the NN used is not a deep-NN, but a single layer NN, that is trained based on the ELM learning algorithm proposed by Huang et al. \cite{ELM}.\\
The results presented in this manuscript show that the proposed PI method can solve several types of exact problems involving parametric DEs with high accuracy and low computational time. For linear and non-linear ordinary differential equations (ODEs) and systems of ODEs (SODEs) the proposed framework achieves machine level accuracy in milliseconds. This makes the X-TFC method is as accurate or more accurate than all other state-of-the-art methods except the classic TFC; although it should be noted that in all cases the classic TFC and X-TFC have solutions errors that are on the same order of magnitude. For linear and non-linear bi-dimensional PDEs, the results achieved by the X-TFC framework are comparable with state-of-the-art methods in terms of speed and outperforms state-of-the-art methods in terms of accuracy by four to 11 orders of magnitude. Furthermore, X-TFC was tested on higher dimensional problems such as a 2D time-dependent non-linear PDE and a 3D time-dependent non-linear PDE. These problems showed that while the method's accuracy and computational time are affected by the increase in the number of dimensions, the method is still well suited for a variety of applications involving PDEs of this dimensionality.\\
In fact, X-TFC is well suited for a variety of applications, even real-time applications that require accurate and fast solutions, such as optimal control problems in aerospace applications. To this end, the authors are currently applying X-TFC to optimal control problems such as energy optimal landing on planetary bodies, minimum time orbit transfer, and maximum radius orbit transfer, to name a few. Moreover, X-TFC is being applied to the solution of transport theory problems such as Radiative Transfer problems (for remote sensing applications and space exploration), and Rarefied Gas Dynamic problems. As mentioned previously, the physics-informed X-TFC method can also be used for data-driven solutions and data-driven discovery of parametric DEs. Currently, progress is begin made on the data-driven discovery of problems involving parametric ODEs from both a deterministic and probabilistic approach.\\
Of course, there is still room to further improve the performance and capability of this new PI method.
The authors are currently investigating the possibility to use new activation functions in addition to the well known logistic, hyperbolic tangent, sinusoid, and Gaussian.
Currently under investigation is the use to different probability distribution to sample input weights and biases, in addition to the uniform and normal distributions.
Future work will attempt to create a Distributed-X-TFC framework similar to the one proposed in Ref. \cite{pielm}, to target problems where the latent solution contains piece-wise continuous behaviour, sharp derivatives, and/or the problems are sufficiently complex such as Navier-Stokes type problems.
In addition to exploring these areas further, future work will also focus on the extension of X-TFC to multi-dimensional problems with non-rectangular domains \cite{raissi,pielm,berg}. 
\section*{Conflicts of Interest}
The authors declare no conflict of interest. 
\section*{Acknowledgements}
This work was partially supported by a NASA Space Technology Research Fellowship, Leake [NSTRF 2019] Grant \#: 80NSSC19K1152 and Johnston [NSTRF 2019] Grant \#: 80NSSC19K1149.

In addition, the authors would like to acknowledge Professor Barry D. Ganapol and Mr. Andrea D'Ambrosio for their precious advises that helped to improve this manuscript, and for suggesting a few remote sensing and aerospace applications, where X-TFC can be tested, as future work topics.
\bibliographystyle{ieeetr}
\bibliography{Refs}

\begin{thebibliography}{10}

\bibitem{TFC}
D.~Mortari, ``{The Theory of Connections: Connecting Points},'' {\em MDPI
  Mathematics}, vol.~5, no.~57, 2017.

\bibitem{LDE}
D.~Mortari, ``{Least-squares Solution of Linear Differential Equations},'' {\em
  MDPI Mathematics}, vol.~5, no.~48, pp.~1--18, 2017.

\bibitem{NDE}
D.~Mortari, H.~Johnston, and L.~Smith, ``High accuracy least-squares solutions
  of nonlinear differential equations,'' {\em Journal of Computational and
  Applied Mathematics}, vol.~352, pp.~293 -- 307, 2019.

\bibitem{raissi}
M.~Raissi, P.~Perdikaris, and G.~E. Karniadakis, ``{Physics-informed neural
  networks: A deep learning framework for solving forward and inverse problems
  involving nonlinear partial differential equations},'' {\em Journal of
  Computational Physics}, vol.~378, pp.~686--707, 2019.

\bibitem{leake}
C.~Leake and D.~Mortari, ``{Deep Theory of Functional Connections: A New Method
  for Estimating the Solutions of Partial Differential Equations},'' {\em
  Machine Learning and Knowledge Extraction}, vol.~2, no.~1, pp.~37--55, 2020.

\bibitem{ELM}
G.-B. Huang, Q.-Y. Zhu, and C.-K. Siew, ``{ Extreme learning machine: Theory
  and applications },'' {\em Neurocomputing}, vol.~70, pp.~489--501, May 2006.

\bibitem{RK}
J.~Dormand and P.~Prince, ``{A Family of Embedded Runge-Kutta Formulae},'' {\em
  J. Comp. Appl. Math.}, vol.~6, pp.~19--26, 1980.

\bibitem{MCPI}
J.~L. Junkins, A.~B. Younes, R.~Woollands, and X.~Bai, ``{Picard Iteration,
  Chebyshev Polynomials, and Chebyshev Picard Methods: Application in
  Astrodynamics},'' {\em \emph{The Journal of the Astronautical Sciences}},
  vol.~60, pp.~623--653, December 2015.

\bibitem{Spectral}
D.~Gottlieb and S.~A. Orszag, {\em {Numerical Analysis of Spectral Methods:
  Theory and Applications}}.
\newblock Society for Industrial and Applied Mathematics, 1977.

\bibitem{NM_OP}
A.~Gil, J.~Segura, and N.~Temme, {\em Numerical Methods for Special Functions}.
\newblock Society for Industrial and Applied Mathematics, 1 2007.

\bibitem{AA}
C.~Lanczos, {\em Applied Analysis}.
\newblock New York: Dover Publications, Inc., 1957.

\bibitem{TFC-Control}
R.~Furfaro and D.~Mortari, ``{Least-squares solution of a class of optimal
  space guidance problems via Theory of Connections},'' {\em Acta
  Astronautica}, 2019.

\bibitem{FOL}
H.~Johnston, E.~Schiassi, R.~Furfaro, and D.~Mortari, ``{Fuel-Efficient Powered
  Descent Guidance on Large Planetary Bodies via Theory of Functional
  Connections},'' {\em The Journal of the Astronautical Sciences}, under
  review.

\bibitem{TFCInt}
H.~Johnston and D.~Mortari, ``{Linear Differential Equations Subject to
  Multivalued, Relative and/or Integral Constraints with Comparisons to
  Chebfun},'' {\em SIAM Journal of Numerical Analysis}, 2018.
\newblock Submitted.

\bibitem{rgd}
M.~De~Florio, E.~Schiassi, R.~Furfaro, and B.~D. Ganapol, ``{An Accurate
  Solution for Poiseuille Flow in a Plane Channel via Theory of Functional
  Connections},'' {\em In preparation}, 2020.

\bibitem{barichello}
L.~Barichello and C.~Siewert, ``{A discrete-ordinates solution for Poiseuille
  flow in a plane channel},'' {\em Zeitschrift f{\"u}r angewandte Mathematik
  und Physik ZAMP}, vol.~50, no.~6, pp.~972--981, 1999.

\bibitem{bari}
L.~Barichello, M.~Camargo, P.~Rodrigues, and C.~Siewert, ``{Unified solutions
  to classical flow problems based on the BGK model},'' {\em Zeitschrift
  f{\"u}r angewandte Mathematik und Physik ZAMP}, vol.~52, no.~3, pp.~517--534,
  2001.

\bibitem{ganapol}
B.~D. Ganapol, ``Poiseuille channel flow by adding and doubling,'' in {\em AIP
  Conference Proceedings}, vol.~1786, p.~070009, AIP Publishing LLC, 2016.

\bibitem{yang}
Y.~Yang, M.~Hou, and J.~Luo, ``{A novel improved extreme learning machine
  algorithm in solving ordinary differential equations by Legendre neural
  network methods},'' {\em Advances in Difference Equations}, vol.~2018, no.~1,
  p.~469, 2018.

\bibitem{FEM}
J.~N. Reddy, ``{An Introduction to the Finite Element Method},'' {\em Journal
  of Pressure Vessel Technology}, vol.~111, pp.~348--349, 08 1989.

\bibitem{FEA1}
J.~Argyris and S.~Kelsey, ``{Energy Theorems and Structural Analysis: A
  Generalized Discourse with Applications on Energy Principles of Structural
  Analysis Including the Effects of Temperature and Non‐Linear
  Stress‐Strain Relations},'' {\em Aircraft Engineering and Aerospace
  Technology}, vol.~26, no.~10, pp.~347--356, 1954.

\bibitem{FEA2}
M.~J. Turner, R.~W. Clough, H.~C. Martin, and L.~J. Topp, ``{Stiffness and
  Deflection Analysis of Complex Structures},'' {\em Journal of the
  Aeronautical Sciences}, vol.~23, pp.~805--823, sep 1956.

\bibitem{FEA3}
R.~W. Clough, {\em {The finite element method in plane stress analysis}}.
\newblock American Society of Civil Engineers, 1960.

\bibitem{OrigOdePde}
I.~E. Lagaris, A.~Likas, and D.~I. Fotiadis, ``{Artificial neural networks for
  solving ordinary and partial differential equations},'' {\em IEEE
  Transactions on Neural Networks}, vol.~9, pp.~987--1000, Sept 1998.

\bibitem{ModernPDE}
J.~Sirignano and K.~Spiliopoulos, ``{\text{DGM}: A deep learning algorithm for
  solving partial differential equations},'' September 2018.

\bibitem{pielm}
D.~Vikas and S.~Balaji, ``{Physics-Informed Extreme Learning Machine (PIELM)- A
  Rapid Method For The Numerical Solution Of Partial Differential Equations},''
  {\em arXiv}, vol.~Xiv:1907.03507v1, 2019.

\bibitem{CoonsPatch}
S.~A. Coons, ``{SURFACES FOR COMPUTER-AIDED DESIGN OF SPACE FORMS},'' tech.
  rep., Massachusetts Institute of Technology, Cambridge, MA, USA, 1967.

\bibitem{M-TFC}
D.~Mortari and C.~Leake, ``{The Multivariate Theory of Connections},'' {\em
  MDPI Mathematics}, vol.~7, no.~3, p.~296, 2019.

\bibitem{M-TFC-PDE}
C.~Leake and D.~Mortari, ``{An Explanation and Implementation of Multivariate
  Theory of Connections via Examples},'' in {\em 2019 AAS/AIAA Astrodynamics
  Specialist Conference, Portland, MN, August 11--15, 2019}, AAS/AIAA, 2019.

\bibitem{SVM-TFC}
C.~Leake, H.~Johnston, L.~Smith, and D.~Mortari, ``{Analytically Embedding
  Differential Equation Constraints into Least Squares Support Vector Machines
  Using the Theory of Functional Connections},'' {\em Machine Learning and
  Knowledge Extraction}, vol.~1, pp.~1058--1083, Oct. 2019.

\bibitem{glam}
E.~Schiassi, R.~Furfaro, J.~S. Kargel, C.~S. Watson, D.~H. Shugar, and U.~K.
  Haritashya, ``{GLAM Bio-Lith RT: A Tool for Remote Sensing Reflectance
  Simulation and Water Components Concentration Retrieval in Glacial Lakes},''
  {\em Frontiers in Earth Science}, vol.~7, 2019.

\bibitem{schiassi}
E.~Schiassi, R.~Furfaro, and D.~Mostacci, ``{Bayesian inversion of coupled
  radiative and heat transfer models for asteroid regoliths and lakes},'' {\em
  Radiation Effects and Defects in Solids}, vol.~171, no.~9-10, pp.~736--745,
  2016.

\bibitem{hapke81}
B.~Hapke, ``{Bidirectional reflectance spectroscopy: 1. Theory},'' {\em Journal
  of Geophysical Research: Solid Earth}, vol.~86, no.~B4, pp.~3039--3054, 1981.

\bibitem{hapke96}
B.~Hapke, ``{A model of radiative and conductive energy transfer in planetary
  regoliths},'' {\em Journal of Geophysical Research: Planets}, vol.~101,
  no.~E7, pp.~16817--16831, 1996.

\bibitem{hapke2002}
A.~S. Hale and B.~Hapke, ``{A time-dependent model of radiative and conductive
  thermal energy transport in planetary regoliths with applications to the Moon
  and Mercury},'' {\em Icarus}, vol.~156, no.~2, pp.~318--334, 2002.

\bibitem{autograd}
D.~Maclaurin, D.~Duvenaud, M.~Johnson, and J.~Townsend, ``Autograd.''
  \url{https://github.com/HIPS/autograd}, 2013.

\bibitem{autodiff}
A.~G. Baydin, B.~A. Pearlmutter, A.~A. Radul, and J.~M. Siskind, ``{Automatic
  differentiation in machine learning: a survey},'' 2015.

\bibitem{CNN}
S.~Mall and S.~Chakraverty, ``{Single Layer Chebyshev Neural Network Model for
  Solving Elliptic Partial Differential Equations},'' {\em Neural Processing
  Letters}, vol.~45, no.~3, pp.~825--840, 2017.

\bibitem{BNN}
H.~Sun, M.~Hou, Y.~Yang, T.~Zhang, F.~Weng, and F.~Han, ``{Solving Partial
  Differential Equation Based on Bernstein Neural Network and Extreme Learning
  Machine Algorithm},'' {\em Neural Processing Letters}, vol.~50, no.~2,
  pp.~1153--1172, 2019.

\bibitem{berg}
J.~Berg and N.~Kaj, ``{A unified deep artificial neural network approach to
  partial differential equations in complex geometries},'' {\em
  Neurocomputing}, vol.~317, pp.~28--41, 2018.

\end{thebibliography}
\end{document}